%% file: main.tex
\pdfoutput=1
\documentclass[11pt]{article}

\usepackage{acl}
\usepackage{times}
\usepackage{latexsym}
\usepackage[T1]{fontenc}
\usepackage[utf8]{inputenc}
\usepackage{microtype}
\usepackage{inconsolata}
\usepackage{graphicx}
\usepackage{booktabs}
\usepackage{amsmath}
\usepackage{amssymb}
\usepackage[table]{xcolor}
\usepackage{subcaption}
\usepackage{tikz}
\usepackage{pgfplots}
\usepackage{placeins}
\usepackage{longtable}
\usepackage{listings}
\usepackage[most]{tcolorbox}
\usepackage{url}
\usepackage{xurl}

\pgfplotsset{compat=1.18}
\newcommand{\method}{DecomposeR}
\newcommand{\planner}{R_{\mathrm{plan}}}
\newcommand{\answerer}{R_{\mathrm{ans}}}
\newcommand{\rexec}{R_{\mathrm{exec}}}
\newcommand{\rsynth}{R_{\mathrm{synth}}}
\newcommand{\crub}{C_{\mathrm{rub}}}
\newcommand{\csearch}{C_{\mathrm{search}}}
\newcommand{\dsearch}{D_{\mathrm{search}}}
\newcommand{\qsearch}{Q_{\mathrm{search}}}
\newcommand{\egraph}{E_{\mathrm{graph}}}
\newcommand{\bsynth}{B_{\mathrm{synth}}}
\newcommand{\icross}{I_{\mathrm{cross}}}
\newcommand{\bsearch}{B_{\mathrm{search}}}
\newcommand{\kembed}{K_{\mathrm{emb}}}
\newcommand{\klex}{K_{\mathrm{lex}}}
\newcommand{\jrub}{J_{\mathrm{rub}}}
\newcommand{\ubranch}{U_{\mathrm{branch}}}
\newcommand{\graph}{\mathcal{G}}
\newcommand{\nodes}{\mathcal{N}}
\newcommand{\edges}{\mathcal{E}}
\newcommand{\rubrics}{\mathcal{R}}
\newcommand{\searchnodes}{\mathcal{N}_{s}}
\newcommand{\aggnodes}{\mathcal{N}_{a}}
\newcommand{\runinhead}[1]{\noindent\textbf{#1}\ }

\definecolor{ThinkTag}{HTML}{7C3AED}
\definecolor{NodeStartTag}{HTML}{0E7490}
\definecolor{NodeEndTag}{HTML}{B45309}

\lstdefinestyle{appendixprompt}{
  basicstyle=\ttfamily\small,
  breaklines=true,
  breakatwhitespace=false,
  columns=fullflexible,
  keepspaces=true,
  showstringspaces=false,
  tabsize=2,
  literate=
    {<think>}{{{\textcolor{ThinkTag}{\bfseries\textless think\textgreater}}}}7
    {</think>}{{{\textcolor{ThinkTag}{\bfseries\textless/think\textgreater}}}}8
    {<node id=}{{{\textcolor{NodeStartTag}{\bfseries\textless node id=}}}}9
    {</node>}{{{\textcolor{NodeEndTag}{\bfseries\textless/node\textgreater}}}}7
}

\newtcblisting{promptbox}{
  enhanced,
  breakable,
  colback=gray!2,
  colframe=black!45,
  listing only,
  listing options={style=appendixprompt},
  left=1mm,
  right=1mm,
  top=1mm,
  bottom=1mm,
  boxrule=0.4pt,
  arc=1mm
}

\newcommand{\promptfigcaption}[1]{%
  \refstepcounter{figure}%
  \begin{center}
    \small Figure~\thefigure: #1
  \end{center}%
  \vspace{0.5em}%
}

\input{figures/rl_curves_pgf}

\title{\method: Planner-Centric Reinforcement Learning for Deep Research
       with Structure-Aware Reward}

\author{Mustafa Anis Hussain, Xinle Wu, Yao Lu \\
National University of Singapore\\
\texttt{\{mustafah,wuxl,luyao\}@comp.nus.edu.sg} \\
}

\begin{document}
\maketitle

\begin{abstract}
Deep research tasks require LLMs to plan what to investigate, retrieve
evidence, and synthesize long-form answers across multiple branches of
inquiry.
Existing training paradigms either rely on short-form verifiable QA as a
proxy or optimize monolithic long trajectories, which makes planning and
execution difficult to disentangle and yields weak credit assignment for
the planning process.
We propose \method, a planner-centric deep research framework that
represents research plans as typed directed acyclic graphs (DAGs),
allowing planning to be made explicit, structured, and rewardable.
We train a Qwen3-8B model in two stages: planner reinforcement learning (RL) first learns
graph structure and query decomposition to improve research planning, and
answerer reinforcement learning (RL) then learns branch-level execution and final synthesis
conditioned on the learned plan.
By assigning rewards to explicit planner tokens and structured components
rather than to a flat trajectory, \method{} enables finer-grained
optimization of planning while reducing the ambiguity of end-to-end
training.
Experiments show that \method{}-8B improves
over strong comparable open baselines by 5.1–8.0 points on 
popular long-form benchmarks due to improved planning and answering capabilities.
\end{abstract}

\section{Introduction}
\label{sec:intro}

Deep research (DR) requires planning what to investigate, issuing
multiple searches, selecting evidence, and synthesizing findings into a
citation-grounded long-form report. Such workflows are increasingly
important for scientific literature synthesis, clinical evidence
aggregation, and general investigative tasks
\citep{asai2024openscholar,singh2025scholarqa,arora2025healthbench,du2025deepresearchbench}.
Frontier proprietary systems demonstrate the practical value of this
workflow
\citep{openai2025deepresearch,google2025gemini25deepresearch,perplexity2025deepresearch},
while open alternatives either wrap capable LLMs in training-free
retrieval pipelines \citep{li2025webthinker,li2025webweaver} or
post-train smaller models with reinforcement learning
\citep{jin2025searchr1,song2025r1searcher,chen2025research,mei2025o2searcher}.

Despite this progress, current RL-trained DR systems share two
reward-side limitations that bound their training signal. First,
credit assignment is ambiguous. Reasoning, search, evidence
selection, branch synthesis, and final writing are interleaved in a flat
ReAct-style trajectory \citep{yao2023react}, and a single scalar reward
derived from the final answer is propagated across every model action. A
weak final answer therefore cannot be attributed to a specific failure
point so the gradient cannot direct improvement towards the
responsible component. Second, reward is sparse. Intermediate
planning, search, and synthesis decisions receive little to no direct supervision
\citep{jin2025searchr1,song2025r1searcher,chen2025research};
every intermediate action is shaped only through its eventual effect on
the terminal answer, which both slows learning and destabilizes credit
propagation across long DR trajectories.

\begin{figure*}[t]
    \centering
    \includegraphics[width=0.8\linewidth]{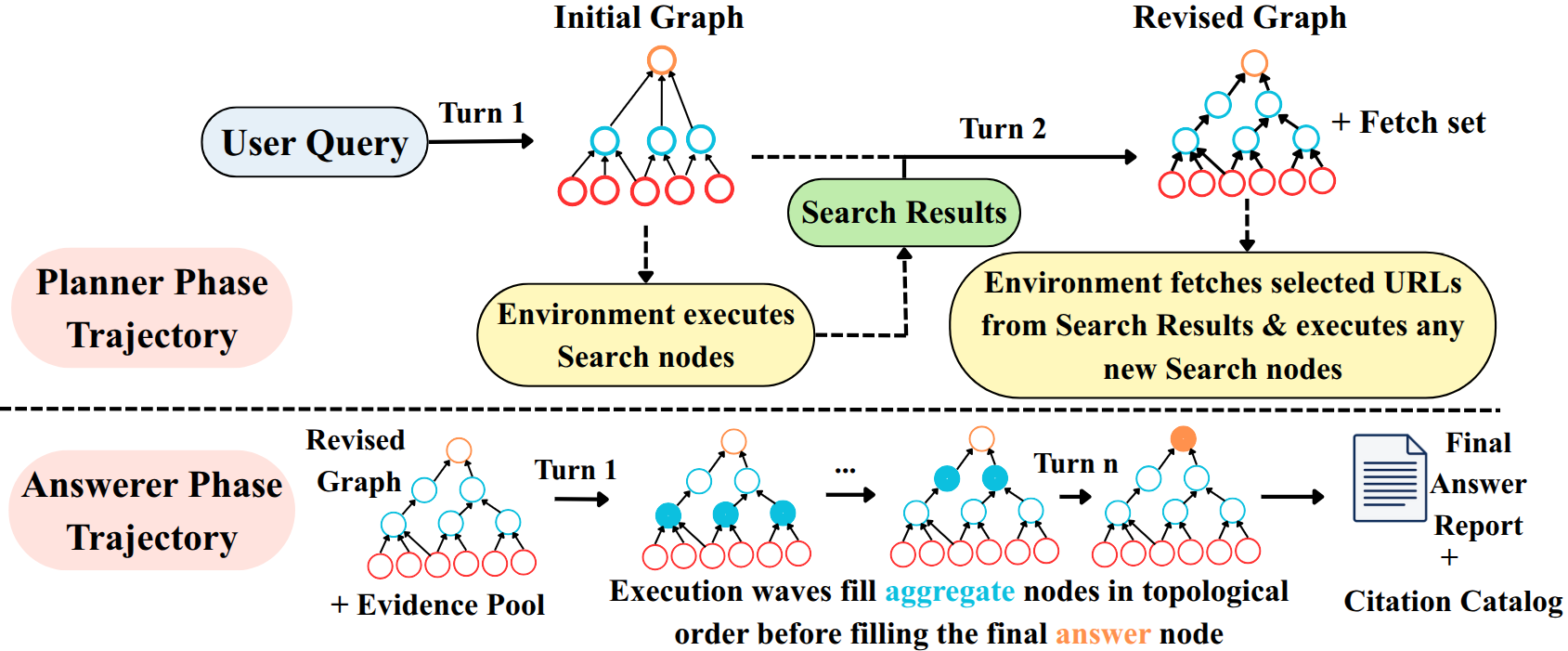}
    \caption{\method{} rollout structure.
    The planner emits an initial typed DAG, receives search results,
    revises the graph and fetch set, and then the answerer executes the
    revised graph in topological waves before writing the final answer.}
    \label{fig:system_overview}
\end{figure*}

We introduce \method, which addresses both limitations through a single
design principle: structure-aware reward modeling for deep
research. The research plan is materialized as an explicit typed DAG
(Figure~\ref{fig:system_overview}) in which search nodes issue web
queries, aggregate nodes synthesize branch-level conclusions, and a
terminal answer node integrates branches into the final report. Because
every plan component is individually addressable, reward can be assigned
to the corresponding part of the research process rather than only to
the final report. Planner rewards score rubric coverage over plan-node
fields, search quality over search nodes, and structural behavior such
as branch breadth and evidence reuse; answerer rewards score
branch-level execution and final synthesis. This design improves credit
assignment by tying reward to the component responsible for each
decision, and reduces sparsity by replacing a single trajectory-end
scalar with signals distributed across the decomposition.

We evaluate \method{} on three long-form benchmarks:
DeepResearchBench, ResearchQA-Mini, and HealthBench. \method{} improves
over strong comparable open baselines by 5.1--8.0 points despite a
substantially smaller end-to-end training budget.

Our contributions are: (1) \textbf{Structure-aware reward modeling for
deep research}: we materialize the DR plan as a typed DAG whose
structural components are directly addressable by reward functions,
allowing separate signals to optimize the corresponding plan and
execution components rather than being read off a trajectory terminus;
(2) \textbf{\method{} framework}: we instantiate these rewards with
typed decomposition graphs and staged reinforcement learning, making
search breadth, evidence reuse, branch integration, execution fidelity,
and final answer quality addressable optimization targets; and
(3) \textbf{Empirical validation}: we evaluate on three long-form
benchmarks, with analysis showing that the reward components, structured
decomposition interface, and search-conditioned revision each contribute
to the final gains.

\section{Related Work}

\runinhead{Deep research agents.}
Recent DR systems use web-scale retrieval and long-form synthesis to
answer open-ended queries.
Frontier systems such as OpenAI Deep Research rely on strong base models
and inference-time orchestration \citep{openai2025deepresearch}.
WebWeaver moves toward more explicit structure through dynamic outlines,
evidence memory, and citation-grounded writing, but remains an
inference-time dual-agent framework rather than an RL objective over a
typed plan \citep{li2025webweaver}.
WebThinker interleaves thinking, searching, navigation, and drafting
inside a web-enabled reasoning loop and improves tool use with online
preference optimization \citep{li2025webthinker}.
WebSailor and WebExplorer focus on long-horizon web agents, constructing
high-uncertainty or evolved web-navigation tasks and training agents to
handle many tool calls \citep{li2025websailor,liu2025webexplorer}.
% DR-Tulu directly trains an open long-form DR model using evolving
% rubrics that adapt to on-policy behavior.
These systems demonstrate the importance of search and synthesis, while
\method{} focuses on making the research plan itself a structured,
separately rewarded policy object.

\runinhead{RL for search and tool use.}
GRPO and RLVR-style post-training have been applied to reasoning,
search, and tool use \citep{shao2024deepseekmath,deepseek2025r1}.
Search-R1 trains models to interleave reasoning and search with
retrieved-token masking and outcome rewards \citep{jin2025searchr1},
while R1-Searcher studies outcome-based RL for invoking search without
process supervision \citep{song2025r1searcher}.
ReSearch similarly treats search calls as part of the reasoning chain
using explicit \texttt{\textless search\textgreater} and
\texttt{\textless result\textgreater} tags, and masks retrieved results
from the policy-gradient loss \citep{chen2025research}.
O$^2$-Searcher targets both open-ended and closed-ended QA through a
local search environment and task-aware rewards \citep{mei2025o2searcher}.
ParallelSearch addresses an orthogonal efficiency problem by rewarding
models for decomposing independent sub-queries that can be searched in
parallel \citep{zhao2025parallelsearch}.
\method{} differs by exposing the full research plan as a typed graph
and assigning reward to graph components before answer synthesis is
trained, rather than treating the search trajectory as a flat sequence of
tool actions.

\runinhead{Query decomposition and retrieval-augmented generation.}
Prior decomposition methods decompose complex questions into subproblems
through prompting or modular inference
\citep{khot2023decomp,wang2023planandsolve,trivedi2023ircot}.
Retrieval-augmented generation systems learn when and what to retrieve
\citep{asai2024selfrag,jiang2023flare}.
\method{} brings these ideas into RL for open-ended DR by making
decomposition a trainable graph policy rather than an unobserved latent
reasoning trace.

% -------------------------------------------------------------------
\section{Methodology}
\label{sec:methodology}

\subsection{Problem Formulation}
\label{sec:problem_formulation}

Existing RL-trained DR systems model the agent rollout as a flat
sequence
\begin{equation}
\label{eq:flat}
\tau_{\mathrm{flat}} = (x, m_1, o_1, \ldots, m_T, o_T, y),
\end{equation}
in which model actions $m_t$ and tool observations $o_t$ interleave
until a final answer $y$ is emitted, and a single scalar reward
$R(x,y)$ is propagated through every $m_t$ for policy update. The
limitations identified in Sec.~\ref{sec:intro} are directly readable
from this formulation. First, every $m_t$ -- whether it carries a
planning decision, a search query, or a synthesis sentence
-- is updated under the same $R$, so the gradient cannot localize the failure to
a specific role. Second, $R(x,y)$ supervises the entire
$\{m_t\}_{t=1}^{T}$, leaving each intermediate $m_t$ shaped only
through its eventual effect on the terminal $y$. Third, $m_t$ is an
unstructured token sequence, so dependencies between subtasks and the
reuse of evidence across branches have no native representation.

\method{} replaces this flat trajectory with a factored one in which
an explicit plan $\graph$ mediates between a planner policy $\pi_P$ and
an answerer policy $\pi_A$:
\begin{equation}
\label{eq:factored}
\tau = (x, \graph_0, Z, \graph_1, F, O_{1:K}, y),
\end{equation}
where $\graph_0$ is an initial plan emitted by $\pi_P$,
$Z = \mathcal{E}(\graph_0)$ are search observations returned by the
environment $\mathcal{E}$, $\graph_1$ is the planner's revised plan,
$F$ are fetched documents, $O_{1:K}$ are intermediate outputs produced
by $\pi_A$ in $K$ topological waves, and $y$ is the final answer. The
joint trajectory distribution factorises as
\begin{equation}
\begin{aligned}
\pi(\tau|x)
&= \pi_P(\graph_0|x)\,\mathcal{E}(Z|\graph_0) \\
&\cdot \pi_P(\graph_1, F|x, \graph_0, Z)\,
   \pi_A(O_{1:K}, y|x, \graph_1, F)
\end{aligned}
\end{equation}
This structure directly resolves the limitations of
Eq.~\ref{eq:flat}: $\pi_P$ and $\pi_A$ become separately optimisable
policies, each amenable to its own reward (resolving role
entanglement); rewards can be assigned to $\graph_1$, to individual
$O_k$, and to $y$ respectively, distributing supervision across the
rollout (resolving sparsity); and $\graph$ is an explicit structured
object that the planner emits and the answerer consumes (resolving
expressiveness, with the structure detailed in
Sec.~\ref{sec:plan_repr}). Consequently the training objective
decomposes into two role-specific objectives,
\begin{equation}
\begin{aligned}
\mathcal{J}_P(\theta_P) &= \mathbb{E}_x\!\left[R_P(\graph_1; x, Z)\right], \\
\mathcal{J}_A(\theta_A) &= \mathbb{E}_x\!\left[R_A(O_{1:K}, y; x, \graph_1)\right]
\end{aligned}
\end{equation}
optimised in two sequential RL stages with $R_P$ and $R_A$ 
% defined in Sec.~\ref{sec:training}.
representing the planner and answerer reward respectively.

\subsection{Overview}
\label{sec:overview}

Figure~\ref{fig:system_overview} illustrates the \method{} rollout,
which is identical at training and inference time up to the
application of gradients. Given a query $x$, the planner first emits an
initial plan $\graph_0$ in the form of a typed DAG (the construction
of which is detailed in Sec.~\ref{sec:plan_repr}). The environment
$\mathcal{E}$ executes every search node in $\graph_0$ and returns the
observation set $Z$. The planner consumes $Z$ and emits a revised plan
$\graph_1$ together with a small set of URLs $F$ to be fetched;
$\mathcal{E}$ then runs only the newly added or modified search nodes
in $\graph_1$ and fetches the URLs in $F$, reusing observations from
unchanged search nodes. Finally, the answerer fills the aggregate
nodes of $\graph_1$ in topological waves and produces the final
citation-grounded report at the answer node. The three design choices
underlying this pipeline are justified below.

\runinhead{Why a typed DAG.}
A typed DAG with search-, aggregate-, and answer-type nodes makes
cross-source evidence reuse and hierarchical synthesis explicit at the
structural level, both of which linear lists and tree-structured plans
cannot natively express. The structure further exposes well-defined
properties-- branch breadth, cross-branch integration, query
distinctness-- as addressable units that the planner reward can target
directly (Sec.~\ref{sec:planner_rl}).

\runinhead{Why two-turn planning.}
$\graph_0$ is generated from $x$ alone and therefore reflects only the
planner's parametric guess at what is worth investigating. Parametric
knowledge frequently misaligns with what is actually retrievable on
the open web: queries can return sparse, off-topic, or outdated
snippets. The second turn closes the planning loop with retrieval
reality, letting the planner drop ineffective subqueries, add new
subqueries inspired by retrieved snippets, and concentrate full-page
fetching on the URLs most likely to support the synthesis.

\runinhead{Why decouple planner and answerer.}
Collapsing planning and answering into one policy under a single
trajectory-level reward conflates their failure modes: a low answer
score cannot distinguish a weak plan executed faithfully from a strong
plan executed poorly. By training the two roles sequentially,
Planner RL optimises rewards that depend only on $\graph_1$ and never
on answerer output, so plan quality is isolated from execution noise.
Answerer RL then trains against a fixed planner, so answerer credit is
not confounded by drifting plan quality. Staging also reduces the
non-stationarity inherent to jointly training two policies that act as
each other's environment, leaving two stable single-policy
optimisation problems.

\subsection{Plan Representation}
\label{sec:plan_repr}

The plan $\graph$ must satisfy three requirements: expose
addressable structure for fine-grained reward design, capture
cross-source evidence reuse and hierarchical synthesis, and admit
efficient execution. Linear subquery lists provide only a flat
sequence with no notion of branch reuse, and tree-structured subgoal
plans forbid a node from being shared across multiple descendants. We therefore represent $\graph$ as a typed directed acyclic
graph, whose construction and execution we now detail.

\runinhead{Node types.}
$\graph$ contains three node types. A search node $s \in \searchnodes$
carries a query string and a list of key points specifying what
evidence the query should return. An aggregate node
$a \in \aggnodes$ carries a synthesis brief consisting of a need
statement and key points to be covered in its output. The terminal
answer node $v_{\mathrm{ans}}$ carries the final synthesis brief; it
has no outgoing edges, and exactly one such node exists per plan.

\runinhead{Edges and validity.}
A directed edge $u \to v$ declares that the output of $u$ is consumed
by $v$: edges from search nodes carry retrieved evidence, edges
between aggregate nodes carry branch-level conclusions, and edges
into $v_{\mathrm{ans}}$ carry the inputs to the final report. A
validator enforces that $\graph$ is a connected DAG with valid JSON
syntax, valid node IDs, allowed node types, no cycles, and exactly one
$v_{\mathrm{ans}}$. Invalid graphs are rejected at the validity gate
before any reward is computed (Sec.~\ref{sec:planner_rl}).

\runinhead{Topological-wave execution.}
Given a valid $\graph_1$, the answerer executes it wave by wave. Wave
$W_\ell$ contains all aggregate nodes whose parents have already been
completed; all nodes in a wave are generated in a single model turn,
exploiting the fact that same-wave aggregates are conditionally
independent given their parents. After all aggregate nodes are filled,
a final turn generates $v_{\mathrm{ans}}$ conditioned on its parent
branches and the cumulative cited evidence. Topological execution
preserves the dependency order implied by $\graph$ while minimising
the number of model turns required.

\subsection{Staged Training}
\label{sec:training}

We train the planner and the answerer sequentially atop a shared
Qwen3-8B backbone with role-conditioned prompts and role-specific LoRA
adapters \citep{hu2022lora}: a planner adapter is active when the
model emits a graph, and an answerer adapter is active when the model
emits aggregate or answer outputs. Training proceeds in three phases:
a cold-start SFT phase that initialises both adapters from teacher
trajectories, a planner RL phase that refines the planner adapter, and
an answerer RL phase that refines the answerer adapter starting from
a checkpoint with the planner-RL adapter merged in. Both RL phases use
Group Relative Policy Optimization (GRPO) \citep{shao2024deepseekmath}
with token-level masking that excludes environment observations from
the policy gradient; the full GRPO objective and masking definitions
are deferred to Appendix~\ref{app:grpo_details}. We describe each phase below,
focusing on the \method{}-specific reward design.

\subsubsection{Cold-Start SFT}
\label{sec:sft}

We first perform supervised fine-tuning to teach the model the basic
graph-emission protocol and wave-execution format, so that
subsequent RL operates on a policy that already produces parseable
outputs rather than wasting samples on protocol violations. A frontier
teacher model (GPT-5-mini) generates complete planner--answerer trajectories for a held-out subset of training queries; each trajectory comprises planner
turns (initial and revised graphs) and answerer turns (aggregate waves
and final answer). We retain only trajectories that pass the graph
validator and the answerer parser end-to-end, discarding those with
malformed JSON, invalid node references, or schema-violating output.
Surviving trajectories yield two streams of turn-level examples--
planner turns under the planner system prompt, and answerer turns
under the answerer system prompt-- on which the planner and answerer
adapters are jointly initialised.

\subsubsection{Planner RL}
\label{sec:planner_rl}

The planner adapter is refined with a reward over three plan-quality
dimensions: semantic coverage of the user's information need,
retrieval quality of the search queries, and structural expressiveness
of the DAG, rewarding each independently before combining them:
\begin{equation}
\label{eq:planner_top}
\begin{aligned}
\planner = \mathbf{1}[\mathrm{Valid}(\graph)]\,\bigl(
&\lambda_{\mathrm{cov}}\,\crub
+ \lambda_{\mathrm{search}}\,\qsearch \\
&+ \lambda_{\mathrm{graph}}\,\egraph
\bigr).
\end{aligned}
\end{equation}
The hard validity gate $\mathbf{1}[\mathrm{Valid}(\graph)]$ ensures
that protocol correctness is established before quality signals take
effect so invalid graphs receive zero reward regardless of their
semantic content. All $\lambda$ terms in this subsection are scalar
reward weights.

\runinhead{Rubric coverage $\crub$.}
The rubric for each training query is generated in advance by a
frontier LLM conditioned on the query together with retrieved web
context, producing query-specific evaluation criteria
$\rubrics = \{(r_u, w_u)\}$ of the form ``the response should cover
$r_u$''. Rubric coverage measures how well the plan has semantically
anticipated these criteria, by taking, for each rubric item, its best
normalised cosine similarity with any non-answer node field
(\texttt{query}, \texttt{key\_points}, \texttt{need}):
\begin{equation}
\crub =
\frac{
  \sum_{u} w_u
  \max_{n \in \nodes \setminus \{v_{\mathrm{ans}}\}}
  \cos^{+}\!\bigl(\phi(u), \phi(n)\bigr)
}{
  \sum_{u} w_u
},
\end{equation}
where $\phi(\cdot)$ denotes the sentence embedding function and $\cos^{+}(\cdot,\cdot)\equiv\max(0,\cos(\cdot,\cdot))$ clamps
negative similarities to zero. Anchoring the rubric to plan-node
fields rather than to the final answer gives the planner direct
semantic feedback on the emitted plan.

\runinhead{Search quality $\qsearch$.}
Search quality measures whether the search nodes target the right
aspects and avoid redundancy:
\begin{equation}
\qsearch =
\lambda_{\mathrm{search,cov}}\,\csearch
+ \lambda_{\mathrm{search,dist}}\,\dsearch ,
\end{equation}
where $\csearch$ is the rubric-to-search-query similarity ($\crub$
restricted to search nodes), and
$\dsearch = 1 - \bar{s}_{\mathrm{pair}}$ is the complement of the mean
pairwise cosine similarity among search queries. Thus $\csearch$
rewards on-topic retrieval, while $\dsearch$ rewards nonredundant
queries.

\runinhead{Graph expressiveness $\egraph$.}
Graph expressiveness combines synthesis breadth $\bsynth$,
cross-branch integration $\icross$, and search breadth $\bsearch$:
\begin{equation}
\label{eq:egraph}
\begin{aligned}
\egraph =
&\lambda_{\mathrm{graph,synth}}\,\bsynth
+ \lambda_{\mathrm{graph,cross}}\,\icross \\
&+ \lambda_{\mathrm{graph,search}}\,\bsearch .
\end{aligned}
\end{equation}
The subterms are
\begin{align}
\bsynth &= f\!\bigl(|\mathrm{pa}_{\aggnodes}(v_{\mathrm{ans}})|\bigr)
           \cdot d_{\mathrm{par}}, \\
\icross &= \tfrac{1}{2}\bigl(r_{\mathrm{fan}} + r_{\mathrm{int}}\bigr), \\
\bsearch &= f\!\bigl(|\searchnodes|\bigr),
\end{align}
where $f(k)=1-1/k$ is a diminishing-returns count score,
$d_{\mathrm{par}}$ is the mean pairwise cosine diversity among the
answer node's aggregate parents, $r_{\mathrm{fan}}$ is the fraction of
search nodes feeding at least two aggregate nodes, and
$r_{\mathrm{int}}$ is the fraction of aggregate nodes integrating at
least two parent inputs. Thus $\bsynth$ rewards diverse direct inputs
to the final answer, $\icross$ rewards evidence reuse and multi-source
branch integration, and $\bsearch$ rewards search breadth.

\begin{figure}[t]
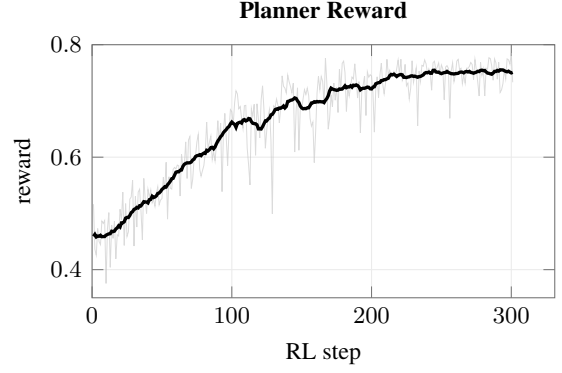

\centering
\PlannerRewardCurve
\caption{Planner RL reward dynamics. The black curve shows the smoothed
planner reward and the grey trace shows raw rollout means. Structural
diagnostics are reported in Appendix~\ref{app:training_diagnostics}.}
\label{fig:planner_training}
\end{figure}
% \FloatBarrier

\subsubsection{Answerer RL}
\label{sec:answerer_rl}

With the planner adapter fixed, the answerer adapter is refined to
execute the plan faithfully and synthesise a high-quality final
report. Graphs are generated by the planner-RL model and rolled out
through the environment in advance, so the answerer trains on a
stable distribution of plans rather than on a moving target. We
decompose answerer quality into two dimensions: branch-level
execution fidelity (measuring whether each aggregate node covers its
declared key points) and global synthesis quality (measuring whether
the final answer integrates the branches and satisfies the
rubric) and combine them as
\begin{equation}
\label{eq:ans_top}
\answerer =
\lambda_{\mathrm{exec}}\,\rexec
+ \lambda_{\mathrm{synth}}\,\rsynth
\end{equation}
All $\lambda$ terms in this subsection are scalar
reward weights.
Parse failures receive partial credit computed over the successfully
completed waves; this eliminates the binary reward cliff that
destabilised early experiments, in which a single malformed node would
zero out a trajectory whose mean baseline was well above zero.

\runinhead{Execution reward $\rexec$.}
Execution reward scores each aggregate node independently against its
declared key points. Only aggregate nodes are scored, since search
nodes are executed by $\mathcal{E}$ rather than by the policy. For
each $v \in \aggnodes$, we combine an embedding-level and a lexical
key-point coverage signal:
\begin{equation}
r_v =
\lambda_{\mathrm{exec,emb}}\,\kembed(v)
+ \lambda_{\mathrm{exec,lex}}\,\klex(v)
\end{equation}
$\kembed(v)$ is a greedy bipartite matching score between the node's
\texttt{key\_points} embeddings and its output-sentence embeddings,
weighted by recall; $\klex(v)$ is the mean fraction of each key
point's content tokens present in the output. The embedding signal
captures semantic coverage, while the lexical signal grounds it
deterministically.
Per-node scores are aggregated with a shifted harmonic mean:
\begin{equation}
\rexec =
\frac{|\aggnodes|}{\sum_{v \in \aggnodes} \frac{1}{1 + r_v}} - 1
\in [0,1]
\end{equation}
The shifted harmonic mean maps all-zero to $0$ and all-one to $1$.

% , and
% lets a single weak branch pull down the aggregate without producing
% the discontinuous reward cliff of a min operator.

\runinhead{Synthesis reward $\rsynth$.}
Synthesis reward targets the final answer rather than individual
branches:
\begin{equation}
\rsynth =
\lambda_{\mathrm{synth,judge}}\,\jrub
+ \lambda_{\mathrm{synth,util}}\,\ubranch .
\end{equation}
$\jrub$ is a rubric-weighted judge score from an external LLM on a
five-point scale. $\ubranch$ is the mean
fraction of each branch's content tokens that appear in the final
answer; it provides a deterministic signal that the model is actually
consuming its intermediate branch outputs.

\begin{figure}[t]
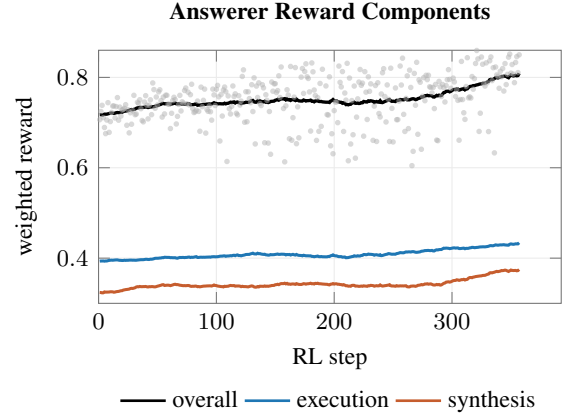

\centering
\AnswererRewardCurve
\caption{Answerer RL reward dynamics. The plot shows overall answerer
reward together with its execution and synthesis contributions. Length
diagnostics are reported in Appendix~\ref{app:training_diagnostics}.}
\label{fig:answerer_training}
\end{figure}
% \FloatBarrier

% -------------------------------------------------------------------
\section{Experiments}
\label{sec:experiments}

\subsection{Experimental Setup}
\label{sec:setup}

\runinhead{Benchmarks and metrics.}
We evaluate \method{} on three long-form deep-research benchmarks
spanning different domains. \textbf{DeepResearchBench (DRBench)}
provides expert-written multi-domain research tasks that require web
exploration, citation-backed synthesis, and report-quality judgment.
\textbf{ResearchQA-Mini (ResQA-Mini)} evaluates scholarly question
answering with survey-derived research questions and rubric items.
\textbf{HealthBench} evaluates responses to health-related
conversations against physician-informed rubric criteria, testing
evidence synthesis in a high-stakes domain. For all three benchmarks
we report the official rubric- or judge-based score; each metric
captures coverage, factual correctness, and response quality on a
scale where higher is better. Appendix~\ref{app:benchmark_eval_details}
provides benchmark-specific evaluation details.

\runinhead{Base model.}
We build \method{} on Qwen3-8B \citep{qwen3} with two
LoRA adapters \citep{hu2022lora}: one for the planner role and one
for the answerer role. All ablation variants use this same backbone
with the same SFT initialization, so that comparisons across variants
isolate the effect of the studied design choice.

\runinhead{Baselines.}
We compare \method{}-8B against two broad types of systems:
\emph{(i) Commercial deep-research systems}: OpenAI (o4-mini) Deep
Research, Gemini Deep Research and Perplexity Deep Research,
accessed through their official APIs. \emph{(ii) Open
deep-research and search agents}:
% DR-Tulu-8B,
WebThinker (QwQ-32B and
32B-DPO), Tongyi DeepResearch-30B, WebExplorer-8B, Search-R1, ASearcher-Web-7B and Qwen3-8B equipped with search. We additionally report a \method{}-8B (SFT only)
variant to isolate the effect of the two RL stages.

\runinhead{Training data.}
We obtain a filtered pool of roughly 4{,}000 long-form English prompts/queries across OpenScholar, SearchArena, and ScholarQA sources. We use 2{,}325 queries for cold-start SFT, converted from teacher planner--answerer trajectories into 10{,}532 turn-level supervision examples. For RL, both
planner RL and answerer RL use the same 1{,}432-query prepared set,
with planner graphs, search results, fetched pages, rubrics, and
citation catalogs attached to each query artifact.

\runinhead{Training protocol.}
Both RL stages use GRPO \citep{shao2024deepseekmath} with verl
\citep{sheng2024verl} and vLLM \citep{kwon2023vllm} for training and rollout generation respectively. 
% Planner RL uses
% learning rate $4\times10^{-6}$, group size 6, training batch size 8,
% and KL coefficient $\beta=0.001$; answerer RL uses learning rate
% $3\times10^{-6}$, group size 4, training batch size 2, and
% $\beta=0.005$. 
The environment uses Jina Search for search-node
execution (up to five results per node) and Serper for URL fetching
(up to four URLs per revision turn). Full GRPO details,
hyperparameters, and prompt templates are in
Appendix~\ref{app:training_details}. Reported numbers are averaged
over multiple random seeds. A central design decision is that \textbf{reward weights are specified
once and kept fixed} across the main experiment and every ablation
variant reported in this section. The reward specification uses no benchmark-specific weight calibration.
Consequently, performance differences across variants reflect the
design choices being studied rather than reward retuning. Reward weights used are reported in full in Appendix~\ref{app:reward_weights}.

\subsection{Main Results}
\label{sec:main_results}

% Table~\ref{tab:main} reports overall performance. \method{}-8B
% (SFT+RL) achieves 41.8 / 42.0 / 71.4 on DRBench, HealthBench, and
% ResQA-Mini, improving over the strongest comparable open baselines by
% 5.1--8.3 points. It matches or exceeds substantially larger
% end-to-end-trained open systems such as Tongyi DeepResearch-30B
% despite a smaller backbone. The gap between \method{}-8B (SFT only)
% and \method{}-8B (SFT+RL) of 2.9--7.6 points across benchmarks
% confirms that staged RL provides substantial gains beyond SFT
% initialization.

\runinhead{Overall performance.}
Table~\ref{tab:main} shows that \method{}-8B (SFT+RL) is the strongest
open model across the three long-form benchmarks compared against models of similar size, achieving 41.8 on DRBench, 42.0 on HealthBench, and 71.4 on
ResearchQA-Mini. Compared with Qwen3-8B + Search, \method{} improves by
7.4 points on DRBench, 22.2 points on HealthBench, and 14.4 points on
ResearchQA-Mini, despite using the same backbone scale. It also
substantially outperforms WebExplorer-8B and other RL-trained search agents
such as Search-R1-7B and ASearcher-Web-7B indicating that
the improvement is not simply due to giving an 8B model access to
search, but training it to organize retrieval and synthesis through
the DecomposeR framework.

\runinhead{RL improves substantially over SFT.}
The SFT-only model already benefits from learning the typed graph and
wave-execution protocol, but the two RL stages provide large additional
gains. The improvement is largest on ResearchQA-Mini, where
rubric-style coverage and synthesis are central, and remains positive on
HealthBench, where domain sensitivity makes unsupported synthesis more
costly.  These gains suggest that SFT is sufficient to initialize the
graph-emission and wave-execution protocol, but that staged RL is needed
for the planner and answerer to use this protocol effectively
.

\runinhead{Comparison with larger open systems.}
Among open deep-research systems, \method{} is the best on average and is competitive with or
stronger than substantially larger models on individual benchmarks. It outperforms Tongyi DeepResearch-30B-A3B on DRBench and ResearchQA-Mini and exceeds
WebThinker-32B-DPO on DRBench and HealthBench, while remaining close on
ResearchQA-Mini. This is notable because those systems rely on much larger
backbones and/or more carefully-engineered inference behavior.

% \begin{table*}[t]
% \centering
% \small
% \begin{tabular}{llccc}
% \toprule
% System & Family & DRBench & HealthBench & ResearchQA-Mini \\
% \midrule
% \multicolumn{5}{l}{\emph{Commercial deep-research systems}} \\
% OpenAI Deep Research & DR agent & 46.9 & 54.0 & 79.1 \\
% Gemini Deep Research & DR agent & 48.9 & 52.5 & 68.8 \\
% Perplexity Deep Research & DR agent & 42.7 & -- & 75.1 \\
% GPT-5 + Search & LLM+search & 50.7 & -- & 78.2 \\
% \midrule
% \multicolumn{5}{l}{\emph{Open deep-research or search agents}} \\
% % DR-Tulu-8B & Open DR & 43.3 & 52.8 & 75.9 \\
% WebThinker QwQ-32B & Open DR & 37.9 & 36.5 & 72.3 \\
% WebThinker-32B-DPO & Open DR & 40.6 & 39.4 & 74.1 \\
% Tongyi DeepResearch-30B-A3B & Open DR & 40.6 & 45.9 & 66.7 \\
% \addlinespace[0.5em]

% WebExplorer-8B & Open search & 36.7 & 33.7 & 65.5 \\
% Search-R1-7B & Open search & 9.7 & 1.1 & 28.4 \\
% ASearcher-Web-7B & Open Search & 10.0 & -13.0 & 22.5 \\
% Qwen3-8B + Search & LLM+search & 34.4 & 19.8 & 57.0 \\
% % \midrule
% \addlinespace[0.5em]
% \textbf{\method{}-8B (SFT only)} & Ours & 36.3 & 39.1 & 63.8 \\
% \textbf{\method{}-8B (SFT + RL)} & Ours & \textbf{41.8} & \textbf{42.0} & \textbf{71.4} \\
% \bottomrule
% \end{tabular}
% \caption{Main results on three long-form deep-research benchmarks.
% \method{}-8B (SFT+RL) refers to the full model trained with both
% planner-RL and answerer-RL stages.}
% \label{tab:main}
% \end{table*}

\begin{table*}[t]
\centering
\small
\definecolor{lightgrey}{gray}{0.93}
\begin{tabular}{llcccc}
\toprule
System & Family & DeepResearchBench & HealthBench & ResearchQA-Mini & Average \\
\midrule
\multicolumn{6}{l}{\emph{Commercial deep-research systems}} \\
OpenAI Deep Research & DR agent & 46.9 & 54.0 & 79.1 & 60.0 \\
Gemini Deep Research & DR agent & 48.9 & 52.5 & 68.8 & 56.7 \\
Perplexity Deep Research & DR agent & 42.7 & -- & 75.1 & -- \\
GPT-5 + Search & LLM+search & 50.7 & -- & 78.2 & -- \\
\midrule
\multicolumn{6}{l}{\emph{Open deep-research or search agents}} \\
% DR-Tulu-8B & Open DR & 43.3 & 52.8 & 75.9 & 57.3 \\
WebThinker QwQ-32B & Open DR & 37.9 & 36.5 & 72.3 & 48.9 \\
WebThinker-32B-DPO & Open DR & 40.6 & 39.4 & 74.1 & 51.4 \\
Tongyi DeepResearch-30B-A3B & Open DR & 40.6 & 45.9 & 66.7 & 51.1 \\
\addlinespace[0.5em]
WebExplorer-8B & Open search & 36.7 & 34.0 & 65.5 & 45.4 \\
Search-R1-7B & Open search & 9.7 & 1.1 & 28.4 & 13.1 \\
ASearcher-Web-7B & Open Search & 10.0 & -13.0 & 22.5 & 6.5 \\
Qwen3-8B + Search & LLM+search & 34.4 & 19.8 & 57.0 & 37.1 \\
% \midrule
% \addlinespace[0.5em]
\rowcolor{lightgrey}
\textbf{\method{}-8B (SFT only)} & Ours & 36.3 & 39.1 & 63.8 & 46.4 \\
\rowcolor{lightgrey}
\textbf{\method{}-8B (SFT + RL)} & Ours & \textbf{41.8} & \textbf{42.0} & \textbf{71.4} & \textbf{51.7} \\
\bottomrule
\end{tabular}
\caption{Main results on three long-form deep-research benchmarks.
\method{}-8B (SFT+RL) refers to the full model trained with both
planner-RL and answerer-RL stages. Entries marked ``--'' indicate
that no public result was available or that running the evaluation was
prohibitively expensive.}
\label{tab:main}
\end{table*}

% =====================================================================
\subsection{Staged Training is Necessary}
\label{sec:exp_staged}

The two experiments in this section answer two distinct questions:
whether staged training outperforms joint training (B1), and whether
the improvement actually traces back to better plans rather than to
the answerer compensating for weak plans (B2). Both halves of
Table~\ref{tab:staged} share the same fixed reward weights as the main
experiment.

% \runinhead{B1: Staged vs.\ Joint RL.}
% We construct a Joint RL baseline that shares with \method{} the same
% two-LoRA architecture, the same SFT initialization, and the same
% typed DAG plan representation; the variants differ only in training
% schedule and reward. In Joint RL, both LoRAs are updated
% simultaneously by gradients flowing from a single trajectory-level
% reward computed from the final answer. This setup isolates the effect of role
% decoupling from any change in architecture or plan representation.
% The top half of Table~\ref{tab:staged} shows that staged training
% outperforms joint training by 3.7--4.7 points across benchmarks. The
% gap is largest on DRBench, whose general-domain queries produce the
% longest rollouts and therefore suffer most from sparse trajectory-end
% credit assignment.

\runinhead{B1: Staged vs.\ Joint RL.}
We construct a Joint RL baseline that shares with \method{} the same
Qwen3-8B backbone, separate planner and answerer LoRA adapters, SFT
initialization, and typed DAG plan representation. The variants differ
only in training schedule and reward assignment. In Joint RL, the
planner and answerer adapters are updated simultaneously from a single trajectory-level reward computed from the final answer. This setup isolates the effect of staged, role-specific optimization from changes in model architecture or plan representation. The top half of
Table~\ref{tab:staged} shows that staged training outperforms joint
training by 3.7--4.7 points across benchmarks. The gap is largest on
DRBench, whose general-domain queries produce the longest rollouts and
therefore suffer most from sparse trajectory-end credit assignment.

\runinhead{B2: Planner quality with a fixed external answerer.}
A natural concern is that the gains in B1 might come from the
answerer learning to compensate for weak plans rather than from the
planner producing better plans. We address this by holding the
answerer fixed and varying only the planner. Specifically, we use
GPT-4.1 with a standardized execution prompt as an external answerer
and feed it plans from four
planners: GPT-5 with a planning prompt (strong training-free
reference), the SFT-only planner, the Joint-RL planner from B1, and
the \method{} planner. The same external answerer executes all four
plan distributions, so differences in final answer quality reflect
only differences in plan quality. The bottom half of Table~\ref{tab:staged} shows that the \method{}
planner produces plans whose downstream answer quality exceeds the
SFT-only and Joint-RL planners considerably and closely trails or
exceeds the GPT-5 planning baseline despite using a substantially
smaller backbone. Combined with B1, this confirms that the gains from
staged training trace back to the planner, not to answerer
compensation.

\begin{table}[t]
\centering
\small
\setlength{\tabcolsep}{1pt}
\begin{tabular}{lccc}
\toprule
Variant & DRBench & HealthBench & ResQA-Mini \\
\midrule
\multicolumn{4}{l}{\emph{B1: Same architecture, different training schedule}} \\
Joint RL & 37.1 & 38.3 & 66.9 \\
\textbf{Staged RL (Ours)} & \textbf{41.8} & \textbf{42.0} & \textbf{71.4} \\
\midrule
\multicolumn{4}{l}{\emph{B2: Different planners, fixed external answerer (GPT-4.1)}} \\
GPT-5 (prompt only) & 39.5 & 40.3 & 68.5 \\
SFT-only planner & 35.4 & 36.0 & 61.3 \\
Joint-RL planner & 37.5 & 37.3 & 65.5 \\
\textbf{\method{} planner} & \textbf{39.8} & \textbf{39.7} & \textbf{69.4} \\
\bottomrule
\end{tabular}
\caption{Validating staged training. B1 compares RL-pipeline
variants trained on the same SFT model.
% , and plan
% representation, differing only in training schedule. 
B2 isolates
planner quality by holding the answerer fixed and varying only the planner.}
\label{tab:staged}
\end{table}

Additional ablation studies reported in Appendix~\ref{app:reward_component_ablation} and ~\ref{app:plan_structure_ablation} further isolate the design
choices behind these gains. The reward-component ablation tests whether performance
comes from a single dominant signal or varied sources. The plan-representation and planner-revision ablations then
test whether typed DAG structure and search-conditioned revision are
necessary by replacing the DAG with
linear/tree structures and by disabling the post-search revision turn.

% =====================================================================
\subsection{Case Study}
\label{sec:case_study}

Appendix~\ref{app:example_rollout} gives a representative \method{}
rollout, including the pre- and post-revision graphs in
Figures~\ref{fig:appendix_rollout_graph_pre}
and~\ref{fig:appendix_rollout_graph_post}, the turn flow in
Table~\ref{tab:appendix_rollout_turn_flow}, and the revised graph
summary in Table~\ref{tab:appendix_rollout_graph_summary}. The example
highlights framework-specific behavior to look for: the initial
decomposition is revised after search feedback; added or modified
search nodes and fetch requests change the evidence available to the
answerer; search evidence can feed multiple downstream aggregates; and
aggregate outputs are executed in topological waves before being
combined by the terminal answer node. 
% These patterns illustrate how the
% typed DAG exposes intermediate research decisions that can be executed,
% diagnosed, and rewarded separately.

% \section{Discussion}

% \method{} makes planning a first-class optimization target.
% The planner results show that explicit graph rewards move both semantic
% coverage and structural behavior simultaneously.
% The answerer results show that factored execution and synthesis rewards
% expose separable learning dynamics that a single terminal reward would
% conflate.

% The main tradeoff relative to flat-trajectory RL is that the DAG
% representation requires a validity gate and a structured reward.
% However, as shown in Table~\ref{tab:reward_weights}, the planner and
% answerer rewards use only mixture weights and the parameter-free $f(k)$
% function; there are no hand-tuned count targets, similarity band
% thresholds, or duplicate cutoffs in the active reward.

% -------------------------------------------------------------------
\section{Conclusion}

We introduced \method, a planner-centric framework for deep research
that represents plans as typed DAGs and trains planner and answerer
roles in separate RL stages.
By assigning reward to explicit graph components, \method{} improves
credit assignment for the planning process.
The planner RL run shows strong gains in reward, validity, search
breadth, aggregate branching, and cross-branch structure, supporting
the core hypothesis that structured plans are a useful optimization
target for deep research systems.
The answerer RL run confirms that execution fidelity and synthesis
quality are separable learning objectives.

\section{Limitations}

The current system fixes the plan at graph revision time and does not
support iterative replanning after the answerer begins execution.
In principle, partial evidence from early aggregate nodes could inform
later search decisions, but this would require variable-length rollouts
that make GRPO credit assignment substantially harder.
The two-turn bounded planner is a deliberate tradeoff that preserves RL
tractability; extending to adaptive replanning is a natural future
direction.

\method{} also depends on external search and fetched pages. Noisy,
incomplete, or poorly ranked evidence can still propagate into the
graph and final answer, and the current system does not perform a
separate source-faithfulness verification step.

Finally, the reward design uses fixed weights across several planner
and answerer components. We keep these weights fixed for all experiments
to avoid benchmark-specific tuning, but this may not be optimal across
domains. In addition, several reward terms rely on embedding-based
rubric or key-point matching, which can over-credit semantically similar
but incomplete content.

% -------------------------------------------------------------------

% \clearpage
\bibliography{custom}

\clearpage
\onecolumn
\appendix
\section*{Appendix}
\input{appendix_material.tex}

\end{document}

%% file: figures/rl_curves_pgf.tex
\definecolor{RLBlue}{HTML}{1F77B4}
\definecolor{RLTeal}{HTML}{2A9D8F}
\definecolor{RLRust}{HTML}{C65D2E}
\definecolor{RLPurple}{HTML}{7B4FA1}

\pgfplotsset{
  rlplot/.style={
    width=\linewidth,
    height=0.64\linewidth,
    xmin=0,
    grid=major,
    major grid style={black!8},
    axis line style={black!55},
    tick style={black!55},
    tick label style={font=\footnotesize},
    label style={font=\footnotesize},
    title style={font=\small\bfseries},
    legend style={
      font=\tiny,
      draw=none,
      fill=white,
      fill opacity=0.85,
      text opacity=1,
      cells={anchor=west}
    },
    legend image post style={line width=1pt},
    scaled ticks=false,
    every axis plot/.append style={line width=1.15pt}
  }
}

\newcommand{\PlannerRewardCurve}{%
\begin{tikzpicture}
\begin{axis}[
  rlplot,
  title={Planner Reward},
  xlabel={RL step},
  ylabel={reward},
  ymin=0.35,
  ymax=0.80,
]
\addplot[black!25, line width=0.35pt, opacity=0.55, forget plot]
  table[col sep=comma, x=step, y=raw_overall]{figures/rl_curves_data/planner_reward.csv};
\addplot[black, line width=1.25pt]
  table[col sep=comma, x=step, y=overall]{figures/rl_curves_data/planner_reward.csv};
\end{axis}
\end{tikzpicture}%
}

\newcommand{\PlannerGraphSizeCurve}{%
\begin{tikzpicture}
\begin{axis}[
  rlplot,
  title={Graph Size by Step},
  xlabel={RL step},
  ylabel={mean count},
  ymin=0,
  ymax=16.5,
  legend columns=4,
  legend style={at={(0.5,-0.20)}, anchor=north, font=\small, draw=none, fill=white, text opacity=1},
]
\addplot[black, line width=1.15pt]
  table[col sep=comma, x=step, y=total_nodes]{figures/rl_curves_data/planner_graph_sizes.csv};
\addplot[RLBlue]
  table[col sep=comma, x=step, y=search_nodes]{figures/rl_curves_data/planner_graph_sizes.csv};
\addplot[RLTeal]
  table[col sep=comma, x=step, y=aggregate_nodes]{figures/rl_curves_data/planner_graph_sizes.csv};
\addplot[RLRust]
  table[col sep=comma, x=step, y=edges]{figures/rl_curves_data/planner_graph_sizes.csv};
\legend{nodes, search, aggregate, edges}
\end{axis}
\end{tikzpicture}%
}

\newcommand{\AnswererRewardCurve}{%
\begin{tikzpicture}
\begin{axis}[
  rlplot,
  title={Answerer Reward Components},
  xlabel={RL step},
  ylabel={weighted reward},
  ymin=0.30,
  ymax=0.86,
  legend columns=3,
  legend style={at={(0.5,-0.30)}, anchor=north, font=\small, draw=none, fill=white, text opacity=1},
]
\addplot[only marks, mark=*, mark size=0.45pt, black!30, opacity=0.50, forget plot]
  table[col sep=comma, x=step, y=raw_overall]{figures/rl_curves_data/answerer_reward_components.csv};
\addplot[black, line width=1.25pt]
  table[col sep=comma, x=step, y=overall]{figures/rl_curves_data/answerer_reward_components.csv};
\addplot[RLBlue]
  table[col sep=comma, x=step, y=execution]{figures/rl_curves_data/answerer_reward_components.csv};
\addplot[RLRust]
  table[col sep=comma, x=step, y=synthesis]{figures/rl_curves_data/answerer_reward_components.csv};
\legend{overall, execution, synthesis}
\end{axis}
\end{tikzpicture}%
}

\newcommand{\AnswererLengthCurve}{%
\begin{tikzpicture}
\begin{axis}[
  rlplot,
  title={Answerer Response Length},
  xlabel={RL step},
  ylabel={mean tokens},
  ymin=6800,
  ymax=10100,
  ytick={7000,8000,9000,10000},
]
\addplot[RLPurple, line width=1.25pt]
  table[col sep=comma, x=step, y=response_length]{figures/rl_curves_data/answerer_response_length.csv};
\end{axis}
\end{tikzpicture}%
}

%% file: appendix_material.tex
\section{Training Details}
\label{app:training_details}

\subsection{Training Data Sources}
\label{app:training_data_sources}

Table~\ref{tab:training_data_sources} gives the exact source breakdown
for the filtered long-form training pool and the artifacts derived from
it.  The active training sets contain 2{,}325 SFT queries, converted
into 10{,}532 supervised turns, and a shared 1{,}432-query RL set used
by both planner RL and answerer RL. Candidate split and rubric columns
are included to document data construction.

\begin{table}[h]
\centering
% \scriptsize
\setlength{\tabcolsep}{3pt}
\begin{tabular}{lrrrrrr}
\toprule
Source & Pool & SFT candidates & SFT used & RL candidates & Rubrics generated & RL used \\
\midrule
OpenScholar & 2{,}577 & 1{,}639 & 1{,}554 & 938 & 899 & 897 \\
SearchArena & 1{,}039 & 630 & 558 & 409 & 404 & 403 \\
ScholarQA & 384 & 231 & 213 & 153 & 132 & 132 \\
\midrule
Total queries/artifacts & 4{,}000 & 2{,}500 & 2{,}325 & 1{,}500 & 1{,}435 & 1{,}432 \\
\bottomrule
\end{tabular}
\caption{Training data source breakdown. ``SFT used'' counts
planner--answerer teacher trajectories used for cold-start SFT.
``Rubrics generated'' counts generated query-specific rubrics. ``RL used''
counts the shared prepared query set used by both planner RL and
answerer RL.}
\label{tab:training_data_sources}
\end{table}

\subsection{Masked GRPO Objective}
\label{app:grpo_details}

This subsection expands the Group Relative Policy Optimization (GRPO)
training protocol summarized in Section~\ref{sec:setup}
\citep{shao2024deepseekmath}. Search-RL systems such as Search-R1 and
ReSearch adapt GRPO to tool environments by treating retrieved text as
environment output and masking it out of the policy-gradient loss
\citep{jin2025searchr1,chen2025research}; O$_2$-Searcher further uses
GRPO with composite rewards for open-ended search tasks
\citep{mei2025o2searcher}. \method{} follows the same separation
principle, but the model-generated objects are typed planner graphs and
answerer node reports rather than inline search traces.

\paragraph{Masked GRPO objective.}
For each input query or prepared answerer artifact $x$, the old policy
samples a group of $G$ rollouts
$\{\tau_i\}_{i=1}^{G}\sim\pi_{\theta_{\mathrm{old}}}(\cdot|x)$.
Each rollout receives a scalar stage reward $R_i$: $R_i=\planner$ for
planner RL and $R_i=\answerer$ for answerer RL. We use the full clipped
GRPO objective of \citet{shao2024deepseekmath}, with the token average
restricted to the model-generated assistant tokens:
\begin{equation}
\label{eq:app_grpo}
\begin{aligned}
\mathcal{J}_{\mathrm{GRPO}}(\theta)
&=\mathbb{E}\!\left[
\frac{1}{G}\sum_{i=1}^{G}\frac{1}{|\mathcal{M}_i|}
\sum_{t\in\mathcal{M}_i}
\left\{
\min\!\left(\rho_{i,t}\widehat{A}_{i},
\operatorname{clip}(\rho_{i,t},1-\epsilon,1+\epsilon)\widehat{A}_{i}\right)
-\beta\,\widehat{D}_{\mathrm{KL}}^{i,t}
\right\}\right], \\
\rho_{i,t}
&=\frac{\pi_{\theta}(y_{i,t}|h_{i,t})}
{\pi_{\theta_{\mathrm{old}}}(y_{i,t}|h_{i,t})},
\qquad
\widehat{A}_{i}=
\frac{R_i-\frac{1}{G}\sum_{j=1}^{G}R_j}
{\operatorname{std}(\{R_j\}_{j=1}^{G})+\delta}, \\
\widehat{D}_{\mathrm{KL}}^{i,t}
&=
\frac{\pi_{\mathrm{ref}}(y_{i,t}|h_{i,t})}
{\pi_{\theta}(y_{i,t}|h_{i,t})}
-\log\!\left(
\frac{\pi_{\mathrm{ref}}(y_{i,t}|h_{i,t})}
{\pi_{\theta}(y_{i,t}|h_{i,t})}
\right)-1 .
\end{aligned}
\end{equation}
Here $h_{i,t}$ is the full dialogue context before token $y_{i,t}$,
$\epsilon$ is the PPO-style clip range, $\beta$ is the KL coefficient,
and $\delta$ is a small numerical constant. The term
$\widehat{D}_{\mathrm{KL}}^{i,t}$ is the positive KL estimator used by
GRPO; unlike RLHF formulations that add a per-token KL penalty into the
reward, this form regularizes the policy directly in the objective. No
learned critic or value head is used; the group mean and standard
deviation provide the baseline and scale. Since our rewards are
trajectory-level stage rewards, the same normalized advantage is assigned
to every unmasked policy token in the rollout.

The mask $\mathcal{M}_i$ contains only policy-generated assistant
tokens. For planner RL, this includes the planner
\texttt{\textless think\textgreater} tokens, graph JSON, revision JSON,
and fetch-request JSON. For answerer RL, this includes the answerer
\texttt{\textless think\textgreater} tokens and XML
\texttt{\textless node\textgreater} blocks for aggregate waves and the
final answer. System prompts, user payloads, Jina search results, Serper
fetch contents, compact dependency payloads, and citation catalogs are
conditioned on but excluded from the policy-gradient and KL terms. This
is the \method{} analogue of retrieved-token masking in inline
search-agent rollouts.

\subsection{Training Hyper-parameters}
\label{app:training_hparams}

Tables~\ref{tab:sft_hparams} and~\ref{tab:grpo_hparams} report the
cold-start SFT and RL hyper-parameters used in the reported run. All
training stages were run on a single node with 8 x NVIDIA RTX PRO 6000
GPUs, each with 96GB of memory. 

\begin{table}[h]
\centering
\small
\setlength{\tabcolsep}{5pt}
\begin{tabular}{@{}p{0.30\linewidth}p{0.62\linewidth}@{}}
\toprule
Setting & Cold-start SFT \\
\midrule
Maximum sequence length & 16{,}384 tokens \\
Epochs & 2 \\
Learning rate & $1.0\times10^{-5}$ \\
Batching & per-device batch size 1; gradient accumulation 8 \\
Temperature for rollout sampling & 1.0 \\
Optimizer and schedule & fused AdamW, cosine schedule \\
Weight decay & 0.0  \\
Precision and memory & BF16, gradient checkpointing, DeepSpeed ZeRO-2 \\
\bottomrule
\end{tabular}
\caption{Cold-start SFT hyper-parameters.}
\label{tab:sft_hparams}
\end{table}

\begin{table}[h]
\centering
\footnotesize
\setlength{\tabcolsep}{4pt}
\begin{tabular}{@{}p{0.28\linewidth}p{0.31\linewidth}p{0.31\linewidth}@{}}
\toprule
Setting & Planner RL & Answerer RL \\
\midrule
Starting checkpoint & Turn-level SFT merged checkpoint & Planner-RL merged checkpoint \\
Rollout backend & VERL with vLLM rollouts & VERL with vLLM rollouts \\
Model dtype & BF16 & BF16 \\
LoRA rank & 64 & 64 \\
LoRA alpha & 128 & 128 \\
Unique prompts per batch & 8 & 4 \\
Rollouts per prompt (group size) & 6 & 6 \\
Mini-batch size & 4 & 2 \\
Mini-batches per GRPO step & 2 & 2 \\
Inner epochs per batch & 1 & 1 \\
Max prompt length in tokens & 4{,}096 & 8{,}192 \\
Max response length in tokens & 8{,}192 & 16{,}384 \\
Max tokens in packed sequence & 12{,}288 & 28{,}672 \\
Temperature for sampling rollouts & 1.0 & 1.0 \\
Top-$p$ for sampling rollouts & 1.0 & 1.0 \\
KL coefficient $\beta$ & 0.001 & 0.005 \\
Learning rate schedule & constant, no warmup & constant, no warmup \\
Learning rate & $4.0\times10^{-6}$ & $4.0\times10^{-6}$ \\
Weight decay & 0.0 & 0.0 \\
Environment & Jina Search, up to five results per search node; Offline Serper fetch, up to four URLs & Offline prepared planner graphs, search results, fetched pages, and citation catalog \\
\bottomrule
\end{tabular}
\caption{GRPO hyper-parameters used for the two RL stages.}
\label{tab:grpo_hparams}
\end{table}

\subsection{Benchmark Evaluation Details}
\label{app:benchmark_eval_details}

\paragraph{DeepResearchBench.}
We evaluated all 100 DeepResearchBench prompts, consisting of 50
Chinese and 50 English tasks from the official query file. Following
the benchmark protocol, we generated one cited long-form report per
prompt, ran the official RACE and FACT evaluation pipelines separately
for the Chinese, English, and combined outputs. RACE was judged with
the official DeepResearchBench LLM judge using
\texttt{openai/gpt-5.5}, while the auxiliary FACT pipeline used
\texttt{openai/gpt-5.4-mini}; the main results table reports the
combined RACE score.

\paragraph{ResearchQA-Mini.}
We evaluated the full 776-question \texttt{test\_mini} split of
ResearchQA-Mini. The generated reports were exported in the expected
ResearchQA response-map format and scored with the official recommended rubric-coverage
judge used by the evaluation wrapper, \texttt{gpt-4.1-mini}.

\paragraph{HealthBench.}
We evaluated 500 samples sampled from the official HealthBench dataset, including 100 prompts from the hard
subset under a hard-sampling fraction of 20 \%, consistent with the original ratio. Generated completions were
attached back to the original HealthBench rows and graded with the
official \texttt{simple-evals} HealthBench rubric implementation using the recommended
\texttt{gpt-4.1-2025-04-14} as the grader.

\subsection{Planner and Answerer Reward Details}
\label{app:reward_weights}
\label{app:reward_details}

\begin{table}[h]
\centering
\small
\setlength{\tabcolsep}{5pt}
\begin{tabular}{lll}
\toprule
Symbol & Description & Value \\
\midrule
\multicolumn{3}{l}{\textit{Planner top-level (Eq.~\ref{eq:planner_top})}} \\
$\lambda_{\mathrm{cov}}$   & Rubric coverage weight   & 0.40 \\
$\lambda_{\mathrm{search}}$  & Search quality weight    & 0.30 \\
$\lambda_{\mathrm{graph}}$ & Graph expressiveness weight & 0.30 \\
\addlinespace[0.3em]
\multicolumn{3}{l}{\textit{Search quality sub-weights}} \\
$\lambda_{\mathrm{search,cov}}$  & Search rubric coverage & 0.50 \\
$\lambda_{\mathrm{search,dist}}$ & Search distinctness    & 0.50 \\
\addlinespace[0.3em]
\multicolumn{3}{l}{\textit{Graph expressiveness sub-weights}} \\
$\lambda_{\mathrm{graph,synth}}$   & Synthesis breadth      & 0.40 \\
$\lambda_{\mathrm{graph,cross}}$ & Cross-branch integration & 0.40 \\
$\lambda_{\mathrm{graph,search}}$  & Search breadth         & 0.20 \\
\addlinespace[0.3em]
\multicolumn{3}{l}{\textit{Answerer top-level (Eq.~\ref{eq:ans_top})}} \\
$\lambda_{\mathrm{exec}}$  & Execution reward weight  & 0.50 \\
$\lambda_{\mathrm{synth}}$ & Synthesis reward weight  & 0.50 \\
\addlinespace[0.3em]
\multicolumn{3}{l}{\textit{Execution sub-weights}} \\
$\lambda_{\mathrm{exec,emb}}$ & Embedding key-point match & 0.50 \\
$\lambda_{\mathrm{exec,lex}}$ & Lexical key-point recall  & 0.50 \\
\addlinespace[0.3em]
\multicolumn{3}{l}{\textit{Synthesis sub-weights}} \\
$\lambda_{\mathrm{synth,judge}}$ & LLM final-answer rubric judge & 0.50 \\
$\lambda_{\mathrm{synth,util}}$  & Deterministic branch-token utilization & 0.50 \\
\bottomrule
\end{tabular}
\caption{All reward weights used in the planner and answerer RL stages.}
\label{tab:reward_weights}
\end{table}
% \FloatBarrier

Table~\ref{tab:reward_weights} separates weights by the object they
score. Planner terms are applied to the revised graph before any final
answer is written, while answerer terms are applied to generated node
reports and the terminal answer under a fixed graph.
Throughout this subsection, $\graph=(\nodes,\edges)$ denotes the
revised planner graph, $\searchnodes\subset\nodes$ denotes its search
nodes, $\aggnodes\subset\nodes$ denotes its aggregate nodes, and
$v_{\mathrm{ans}}$ denotes the terminal answer node. The function
$\phi(\cdot)$ is the sentence-embedding encoder used for reward-side
semantic matching, $\cos(\cdot,\cdot)$ is cosine similarity in that
embedding space, and $\max(0,\cos(\cdot,\cdot))$ clamps negative
similarities to zero.

\paragraph{Planner-stage reward.}
The planner rollout consists of an initial graph, environment search
results, and a planner revision that emits the final graph and fetch
requests. Only the revised graph $\graph$ is rewarded. Invalid graphs
receive zero reward:
\begin{equation}
\label{eq:app_planner_reward}
\planner =
\mathbf{1}[\mathrm{Valid}(\graph)]
\left(
0.40\,\crub + 0.30\,\qsearch + 0.30\,\egraph
\right).
\end{equation}
Here $\mathbf{1}[\mathrm{Valid}(\graph)]$ is one only when the revised
graph parses and satisfies the typed-DAG constraints; otherwise the
planner reward is zero.
The three terms correspond to distinct planner responsibilities:
$\crub$ checks whether the plan creates places for the rubric content
to be answered later, $\qsearch$ checks whether the search nodes expose
useful and non-duplicative retrieval routes, and $\egraph$ checks
whether the graph contains enough reusable synthesis structure for the
answerer to execute.
For a rubric item $u$ with weight $w_u$, let
$\mathrm{fields}(n)$ be the non-empty \texttt{query},
\texttt{key\_points}, and \texttt{need} fields of a non-answer node
$n$. Rubric coverage $\crub$ is the weighted average best match from
each rubric item to any searchable or aggregate plan field:
\begin{equation}
\crub =
\frac{\sum_u w_u
\max_{n\in\nodes\setminus\{v_{\mathrm{ans}}\}}
\max_{c\in\mathrm{fields}(n)}
\max(0,\cos(\phi(u),\phi(c)))}
{\sum_u w_u}.
\end{equation}
Search quality is
\begin{equation}
\qsearch = 0.50\,\csearch + 0.50\,\dsearch ,
\end{equation}
where $\csearch$ is computed like $\crub$ but restricted to search-node
queries. Let $q_i$ be the query string emitted by the $i$th search node
in $\searchnodes$. The distinctness term $\dsearch$ is one minus the
mean clamped cosine similarity over unordered pairs of search queries:
\begin{equation}
\dsearch =
1 -
\frac{2}{|\searchnodes|(|\searchnodes|-1)}
\sum_{i<j}\max(0,\cos(\phi(q_i),\phi(q_j))).
\end{equation}
When there are fewer than two search queries, $\dsearch$ is set to
zero. Thus $\csearch$ rewards query-level recall of important facets,
while $\dsearch$ discourages multiple search nodes from becoming
paraphrases of the same retrieval request.

Graph expressiveness is
\begin{equation}
\egraph =
0.40\,\bsynth + 0.40\,\icross + 0.20\,\bsearch .
\end{equation}
All three subterms use the diminishing-returns function
$f(k)=\max(0,1-1/\max(k,1))$. Let
$P_{\mathrm{ans}}$ be the set of aggregate parents of the answer node
and $d_{\mathrm{par}}$ be one minus the mean pairwise normalized cosine
similarity among their node texts. Then
\begin{align}
\bsynth &= f(|P_{\mathrm{ans}}|)\,d_{\mathrm{par}}, \\
\icross &= \tfrac{1}{2}(r_{\mathrm{fan}}+r_{\mathrm{int}}), \\
\bsearch &= f(|\searchnodes|),
\end{align}
where $\bsynth$ measures how many distinct aggregate branches feed the
answer, $\icross$ measures cross-branch connectivity, and $\bsearch$
measures search breadth. The term $r_{\mathrm{fan}}$ is the fraction of
search nodes with at least two direct downstream children, so it rewards
evidence reuse across branches. The term $r_{\mathrm{int}}$ is the
fraction of aggregate nodes with at least two parent inputs, so it
rewards synthesis nodes that combine evidence or upstream conclusions.
The diminishing-returns function gives most of its gain when moving from
one to two useful branches or search nodes, then saturates as counts
grow; the diversity multiplier $d_{\mathrm{par}}$ prevents duplicate
answer-parent branches from increasing $\bsynth$ cheaply.

\paragraph{Answerer-stage reward.}
The answerer rollout executes the fixed planner graph in topological
aggregate waves and then writes the terminal answer node. For a fully
parsed trajectory,
\begin{equation}
\label{eq:app_answerer_reward}
\answerer = 0.50\,\rexec + 0.50\,\rsynth .
\end{equation}
The execution term is local to aggregate nodes: it asks whether each
branch report fulfilled the key points declared by the planner. The
synthesis term is reserved for the terminal answer, so missing branch
coverage and poor final integration affect different parts of the
reward.
Only aggregate nodes are scored for execution. For each aggregate node
$v$ with key points $P_v$ and generated node text $o_v$,
\begin{equation}
r_v = 0.50\,\kembed(v) + 0.50\,\klex(v).
\end{equation}
A high $r_v$ therefore means that a node report covers its assigned
content targets, not merely that it is fluent or long. Here $P_v$ is
the set of planner-declared key points for aggregate node $v$, and
$o_v$ is the generated XML node text for that aggregate. The embedding
term $\kembed(v)$ is the greedy one-to-one matching score between key
points $p\in P_v$ and output sentences $s$ from $o_v$, normalized by the
number of key points:
\begin{equation}
\kembed(v)=
\frac{1}{|P_v|}
\sum_{(p,s)\in\mathrm{match}(v)}
\max(0,\cos(\phi(p),\phi(s))).
\end{equation}
The lexical term strips inline citations and averages key-point token
recall. Here $\mathrm{tok}(\cdot)$ denotes the set of normalized content
tokens after citation removal:
\begin{equation}
\klex(v)=
\frac{1}{|P_v|}
\sum_{p\in P_v}
\frac{|\mathrm{tok}(p)\cap\mathrm{tok}(o_v)|}
{|\mathrm{tok}(p)|}.
\end{equation}
The embedding term gives credit for semantic coverage of key points,
while the lexical term provides a deterministic guard against high
semantic similarity without concrete key-point content.
Node scores are combined with the shifted harmonic mean
\begin{equation}
\rexec =
\frac{|\aggnodes|}
{\sum_{v\in\aggnodes} \frac{1}{1+r_v}}
-1.
\end{equation}
This preserves sensitivity to weak branches without turning one failed
branch into a total zero reward. We use the shifted harmonic mean rather
than a min operator because early experiments with a hard minimum made
GRPO advantage estimates unstable under small group sizes.

Synthesis reward combines an LLM rubric judge and deterministic branch
utilization:
\begin{equation}
\rsynth = 0.50\,\jrub + 0.50\,\ubranch .
\end{equation}
These two subterms separate external answer quality from structural
fidelity: $\jrub$ measures whether the final report satisfies the
query-level rubric, while $\ubranch$ checks whether the final report
actually uses the branch reports produced during graph execution.
The judge term $\jrub$ is a rubric-weighted average of five-level
scores $\{0,0.25,0.5,0.75,1\}$ returned by the configured judge
(\texttt{gpt-5-mini} in the reported run). The five-level scale is used
because coarser three-level scales merge adjacent rubric-satisfaction
levels, while broader numeric scales introduce more calibration noise
across queries. Branch utilization is
\begin{equation}
\ubranch =
\frac{1}{|B|}
\sum_{b\in B}
\frac{|\mathrm{tok}(b)\cap\mathrm{tok}(y)|}
{|\mathrm{tok}(b)|},
\end{equation}
where $B$ is the set of aggregate branch outputs and $y$ is the final
answer, both after stripping inline citations. Each branch output
$b\in B$ is one completed aggregate-node report. This term is
deterministic and targets the failure mode in which the answerer ignores
its completed branch reports and writes the final answer directly from
parametric knowledge or retrieved evidence.

All reward-side semantic matching uses the
\texttt{sentence-transformers/all-MiniLM-L6-v2} embedding model
\citep{sentenceTransformersAllMiniLML6v2}.

\paragraph{Partial credit for parse failures.}
Planner parse or graph-validity failures receive zero because the
planner reward is defined on a valid typed DAG. Answerer parse failures
are handled differently: if some aggregate waves have already completed,
missing aggregate nodes are assigned zero node scores and the rollout
receives $0.50\,\rexec$ with $\rsynth=0$; if the final answer fails to
parse after all aggregate waves succeed, it receives the same execution
credit with zero synthesis. If no aggregate node is successfully parsed,
the reward remains zero. This change was used because fully zeroing a
long answerer rollout after a late XML parse error produced unstable
GRPO advantages despite useful completed branch reports.

\newpage
\subsection{Training Diagnostics}
\label{app:training_diagnostics}

\begin{figure}[h]
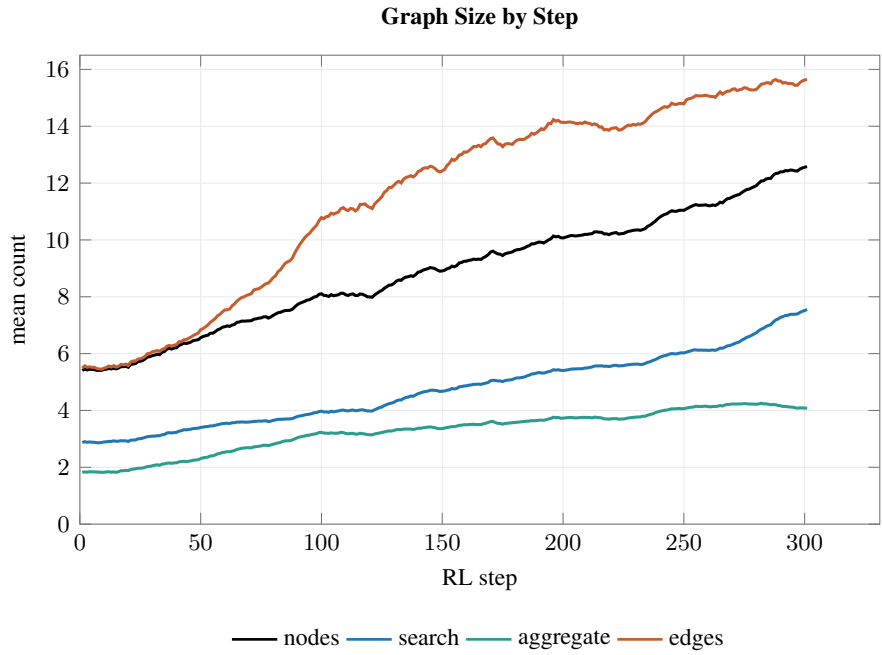

\centering
\begin{minipage}{0.76\linewidth}
\centering
\PlannerGraphSizeCurve
\end{minipage}
\caption{Planner graph size and typed component counts over RL.}
\label{fig:appendix_planner_graph_size}
\end{figure}

\begin{figure}[h]
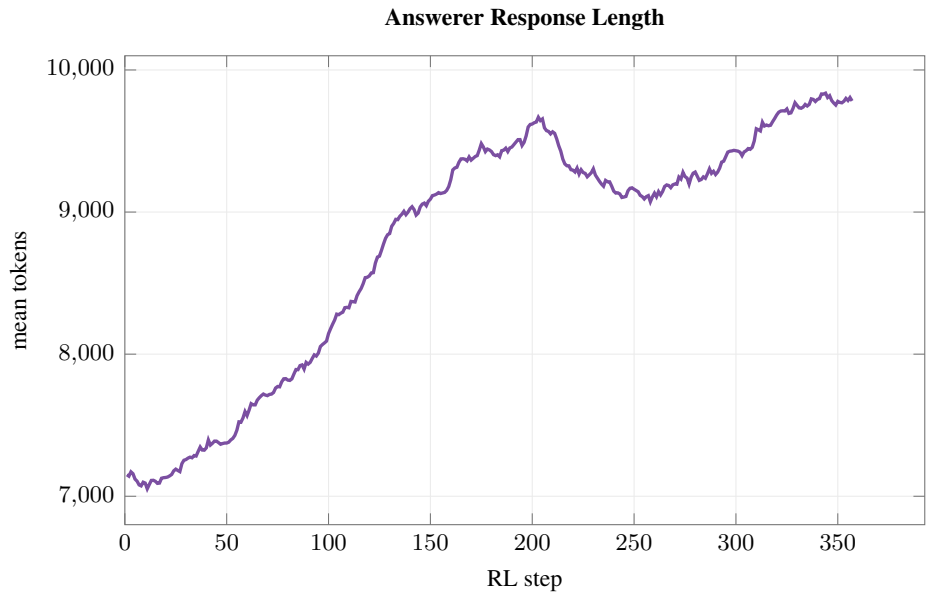

\centering
\begin{minipage}{0.76\linewidth}
\centering
\AnswererLengthCurve
\end{minipage}
\caption{Answerer response length over RL.}
\label{fig:appendix_answerer_length}
\end{figure}

Figures~\ref{fig:appendix_planner_graph_size}
and~\ref{fig:appendix_answerer_length} report the two training
diagnostics omitted from the main text for space. The planner graph
curve shows that planner RL does not merely improve scalar reward: it
also increases typed graph structure, with edge count growing faster
than node count as the model learns to reuse evidence across branches.
The answerer length curve shows that responses become longer during
answerer RL.  We interpret this as a consequence of key-point coverage
pressure rather than length reward hacking, since the active reward has
no direct length term and the reward curve in Figure~\ref{fig:answerer_training}
tracks execution and synthesis improvements.

\newpage

\subsection{Reward Component Ablations}
\label{app:reward_component_ablation}

We ablate the planner and answerer reward components separately,
removing one component at a time from the full reward and retraining
from the SFT checkpoint with the same schedule, data, and fixed
weights on the remaining components. Results are reported in
Table~\ref{tab:appendix_reward_components}.

\paragraph{C1: Planner reward components.}
The top half of Table~\ref{tab:appendix_reward_components} reports
leave-one-out ablations of the three top-level planner reward terms.
Removing rubric coverage $\crub$ causes the largest drop on
ResQA-Mini, whose rubric-based evaluations directly reward semantic
coverage of specific aspects. Removing search quality $\qsearch$
degrades all three benchmarks, falling from 51.7 to 49.1 on average,
confirming that this term shapes retrieval coverage rather than only
acting as a noise term. Removing graph expressiveness $\egraph$ has a
sizeable effect on final answer scores and causes graphs to collapse
toward less integrated, simpler structures. This contrasts with the
full planner-RL diagnostic in Figure~\ref{fig:appendix_planner_graph_size},
where edge count grows faster than node count, indicating increasing
cross-branch reuse.

\paragraph{C2: Answerer reward components.}
The bottom half of Table~\ref{tab:appendix_reward_components} reports
the same ablation for the answerer reward. Removing per-aggregate
execution reward $\rexec$ causes branch-level key-point coverage to
drop substantially, resulting in the largest drop-off across all
answerer sub-components. Removing the rubric judge $\jrub$ reduces
final answer quality on all three benchmarks. The branch utilization
term $\ubranch$ has the smallest standalone effect on benchmark scores,
but when removed the content overlap between the final answer and
intermediate branch outputs drops from 0.82 to 0.61, indicating that
without $\ubranch$ the answerer increasingly ignores its own branch
outputs and re-derives the final report from scratch.

\begin{table}[h]
\centering
\small
\setlength{\tabcolsep}{4pt}
\begin{tabular}{lccc}
\toprule
Variant & DRBench & HealthBench & ResQA-Mini \\
\midrule
\multicolumn{4}{l}{\emph{C1: Planner reward components}} \\
\textbf{Full $\planner$} & \textbf{41.8} & \textbf{42.0} & \textbf{71.4} \\
\quad w/o $\crub$ & 36.3 & 37.9 & 65.2 \\
\quad w/o $\qsearch$ & 38.9 & 39.7 & 68.8 \\
\quad w/o $\egraph$ & 34.1 & 35.8 & 66.3 \\
\midrule
\multicolumn{4}{l}{\emph{C2: Answerer reward components}} \\
\textbf{Full $\answerer$} &  \textbf{41.8} & \textbf{42.0} & \textbf{71.4} \\
\quad w/o $\rexec$ & 36.1 & 31.9 & 62.8 \\
\quad w/o $\jrub$ & 38.0 & 39.3 & 69.8 \\
\quad w/o $\ubranch$ & 40.5 & 41.3 & 70.3 \\
\bottomrule
\end{tabular}
\caption{Leave-one-out ablation of reward components. All variants
keep the fixed reward weights on remaining components unchanged.}
\label{tab:appendix_reward_components}
\end{table}

\subsection{Plan Representation and Revision Ablations}
\label{app:plan_structure_ablation}

This ablation complements the reward-component analysis in
Appendix~\ref{app:reward_component_ablation} by isolating two structural choices in
the planner: whether the plan is a typed DAG rather than a weaker
linear or tree structure, and whether the planner can revise the plan
after seeing search results. Results are reported in
Table~\ref{tab:appendix_plan_structure_ablation}.

\paragraph{D1: Plan structure ablation.}
We replace the typed DAG with two alternative plan representations
while retaining the rest of the \method{} training pipeline.
\textbf{Linear}: the plan consists of search nodes feeding directly
into the answer node with no aggregate layer, approximating the
parallel-subquery structures used by ParallelSearch and SubSearch;
the graph expressiveness reward $\egraph$ is removed because there
is no DAG structure to score, leaving $\crub$ and $\qsearch$ active.
\textbf{Tree}: aggregate nodes are restricted to at most one parent,
approximating hierarchical outline structures of WebWeaver-style
systems; the structural rewards remain active but $r_{\mathrm{int}}$
in $\icross$ is necessarily zero. The top half of
Table~\ref{tab:appendix_plan_structure_ablation} shows that the typed
DAG outperforms both Linear and Tree on every benchmark, with the
largest gap on tasks that require synthesizing evidence across
multiple thematic branches.

\paragraph{D2: Single-turn vs.\ two-turn planning.}
We further ablate the two-turn planner protocol by training a
single-turn variant in which the planner emits $\graph_0$ and
immediately commits to it, with no opportunity to revise after
seeing search results. The bottom half of
Table~\ref{tab:appendix_plan_structure_ablation} shows that two-turn
planning improves over single-turn planning on all three benchmarks,
consistent with the intuition that an initial plan generated from
parametric knowledge alone benefits from being adjusted to what is
actually retrievable.

\begin{table}[h]
\centering
\small
\setlength{\tabcolsep}{3pt}
\begin{tabular}{lccc}
\toprule
Variant & DRBench & HealthBench & ResQA-Mini \\
\midrule
\multicolumn{4}{l}{\emph{D1: Plan structure}} \\
Linear & 34.2 & 31.5 & 57.8 \\
Tree & 38.6 & 39.8 & 65.3 \\
\textbf{Typed DAG (Ours)} & \textbf{41.8} & \textbf{42.0} & \textbf{71.4} \\
\midrule
\multicolumn{4}{l}{\emph{D2: Planner revision}} \\
Single-turn planning & 35.1 & 36.7 & 67.5 \\
\textbf{Two-turn planning (Ours)} & \textbf{41.8} & \textbf{42.0} & \textbf{71.4} \\
\bottomrule
\end{tabular}
\caption{Plan representation ablations. D1 replaces the typed DAG
with a linear (parallel-subquery) or tree (hierarchical outline)
structure. D2 disables the revision turn that follows initial
search.}
\label{tab:appendix_plan_structure_ablation}
\end{table}

\section{Prompt and Rollout Appendix}

This appendix records the exact system prompts used by the \method{}
pipeline and a model-output rollout from the final 8B model. Planner
JSON, evidence payloads, and long answerer text are excerpted to avoid
printing full fetched pages or repetitive answer prose; the graph,
turn flow, execution waves, final answer structure, and citation
catalog are preserved.

\subsection{Execution and SFT Prompts}

\label{app:execution_sft_prompts}

This subsection groups the static system prompts and turn-specific user payloads used for production rollouts and cold-start SFT. The initial planner turn receives only the raw query; after search, the planner receives a structured revision payload. The answerer then receives aggregate-wave payloads and a final-answer payload. The SFT corpus stores these same prompt/completion turns from teacher trajectories.

\paragraph{Planner prompts.}

\begin{promptbox}
You are the planner for a graph-based deep research system.

The same underlying model is used for planning and answering, but this planner role is only responsible for:
1. proposing or revising the research graph
2. choosing which URLs are worth fetching after search results are visible

You do not write the final answer.

Every planner response must follow this pattern:
1. a very brief `<think>` block
2. the required JSON object

Keep the `<think>` block short. One or two sentences is enough. Do not write a long hidden monologue.

Planner graph rules:
- support multiple branches
- allow cross-links when the graph remains a Directec Acyclic Graph (DAG)
- allow aggregate nodes with multiple parents
- allow search nodes that feed more than one aggregate
- allow intermediate aggregate nodes
- end with exactly one terminal `answer` node

Allowed node types:
- `search`
- `aggregate`
- `answer`

Allowed edge types:
- `enables`
- `compares`
- `extends`
- `requires`
- `contextualizes`

Node requirements:
- Every `search` node needs:
  - `id`
  - `type`
  - `need`
  - `query`
- Every `aggregate` node needs:
  - `id`
  - `type`
  - `need`
  - `key_points`
  - `inputs`
- Every `answer` node needs:
  - `id`
  - `type`
  - `need`
  - `inputs`

`key_points` rules for aggregate nodes:
- must be present on every aggregate node
- should contain between 2 to 6 items
- each item should be a concrete content target, not a vague heading
- together they should define what a strong local mini-report must cover

General planning rules:
- search queries should be operationalized subproblems, not paraphrases of the user query
- aggregate nodes should be meaningful mini-reports, not vague buckets
- the final answer node should synthesize across multiple branches instead of restating one branch
- keep the graph acyclic
- do not invent URLs
- do not output any prose outside the `<think>` block and the required JSON
- the edge type should reflect how the search query can help address the aggregate node's needs

Turn 1:
- read the user query
- propose an initial graph
- prefer 4 to 5 search nodes for most deep research queries unless the task clearly needs more

Turn 1 output schema:
<think>...[brief planning note]...</think>
{
  "graph": {
    "nodes": [
      {
        "id": "N1",
        "type": "search",
        "need": "specific information need",
        "query": "concrete search query"
      },
      {
        "id": "N4",
        "type": "aggregate",
        "need": "specific branch report need",
        "key_points": [
          "point one",
          "point two",
          "point three"
        ],
        "inputs": ["N1", "N2"]
      },
      {
        "id": "N7",
        "type": "answer",
        "need": "integrated final answer need",
        "inputs": ["N5", "N6"]
      }
    ],
    "edges": [
      {
        "source": "N1",
        "target": "N4",
        "type": "enables"
      }
    ]
  }
}

Turn 2:
- you will receive the current graph and executed search results
- keep or revise the graph based on the search results and newly-acquired knowledge
- choose the few URLs most worth fetching
- any new search nodes you add in the revised graph will be executed after this turn
- choose ONLY URLs that already appear in the provided search results
- fetch sparingly

Turn 2 revision rules:
- you may revise the structure of the graph in multiple ways 
(e.g. add missing branches if the search evidence reveals a distinct missing sub-problem,
further diverging a single branch or combining multiple branches, adding new search nodes etc.)
- preserve a valid DAG
- keep `key_points` strong and specific after revision

Turn 2 output schema:
<think>...[brief revision note]...</think>
{
  "action": "keep|revise",
  "revised_graph": {
    "nodes": [...],
    "edges": [...]
  },
  "fetch_requests": [
    {
      "source_node_id": "N1",
      "url": "https://example.com/page",
      "reason": "why this page is worth fetching"
    }
  ]
}

Small example of good aggregate `key_points`:
- weak: ["background", "implications"]
- better:
  - "how obesity treatment guidelines changed after 2021"
  - "semaglutide and tirzepatide adoption timeline"
  - "what changed for payer and provider behavior"

Return only:
- one brief `<think>` block
- one JSON object
\end{promptbox}
\promptfigcaption{Planner system prompt}

\begin{promptbox}
{
  "phase": "planner_after_search",
  "query": "{user query}",
  "current_graph": "{initial graph, with search/aggregate/answer nodes}",
  "max_fetches": 4,
  "search_results": "{Jina search results keyed by source search node}",
  "instruction": "Review the evidence. Keep or revise the graph, then choose the few URLs most worth fetching. Choose only from the provided search results. Fetch sparingly. Revise the graph if the current branches under-separate distinct analytical tasks, if any aggregate node is overloaded, or if the evidence suggests a missing branch. If you add new search nodes, they will be executed after this turn. Every aggregate node must keep specific `key_points` with 2-6 items."
}
\end{promptbox}
\promptfigcaption{Planner revision-turn user payload}

\paragraph{Answerer prompts.}

\begin{promptbox}
You are the answerer and executor for a graph-based deep research system.

The planner role is already complete.
You do not revise the graph.
You do not choose fetches.
You do not open new searches.

Your job is to execute the already-planned graph in topological order.

Execution model:
- the environment gives you one execution wave at a time
- each execution wave may contain multiple same-level aggregate nodes
- you must complete every target node in that wave in a single answerer turn
- after upstream aggregates exist, you will later receive a final turn for the terminal answer node

Dependency format:
- execution turns may provide `shared_dependencies` once at the top level instead of repeating them under every target node
- each target node tells you which dependency ids it may use through `dependency_refs`
- if a dependency has `provision: inline`, its evidence is included in the current turn
- if a dependency has `provision: history_reference`, its full content was already provided earlier in the conversation and is not repeated here
- use only the dependency ids referenced for the current target node

Evidence discipline:
- use only the evidence provided or referenced for each target node
- do not borrow facts from sibling nodes unless that content is explicitly included in the target node's dependencies
- if evidence is thin or mixed, say so directly
- do not add unsupported facts

What strong aggregate outputs look like:
- substantial independent deep-research reports, not terse notes
- tightly aligned to the node `need`
- clearly and extensively cover the node `key_points`
- elaborate and synthesize on the provided parents instead of listing them mechanically
- use inline citations for factual claims with the exact evidence ids provided in the prompt, such as `[N3-R2]` or `[N1-F1]`

What strong final answers look like:
- integrate the completed branch reports into a coherent stand-alone report
- preserve the distinctions and conclusions established upstream
- answer the user query directly, not just summarize the graph
- be extended long-form and detailed by default for deep-research queries
- use clear sections and sustained reasoning instead of short summary prose
- preserve and reuse inline evidence citations from the aggregate reports

Every answerer response must follow this pattern:
1. a very brief `<think>` block
2. one or more XML node blocks

Keep the `<think>` block short. One or two sentences is enough.

Execution-turn output format:
<think>brief execution note</think>
<node id="N3">
deep-research report for node N3
</node>
<node id="N4">
deep-research report for node N4
</node>

Final answer output format:
<think>brief synthesis note</think>
<node id="N11">
final integrated deep-research answer
</node>

Rules for XML node blocks:
- emit exactly one `<node ...>` block per required target node
- the `id` must match exactly
- do not wrap the nodes in JSON
- do not add commentary before, between, or after node blocks except the single brief `<think>` block
- do not add a citation bibliography or source block; use inline citations only

Small example for a two-node execution wave:
<think>I should handle the market-structure branch and the governance branch separately, using only their own dependencies.</think>
<node id="N5">
Report for N5...
</node>
<node id="N6">
Report for N6...
</node>

Return only:
- one brief `<think>` block
- the required `<node>` blocks
\end{promptbox}
\promptfigcaption{Answerer system prompt}

\begin{promptbox}
{
  "phase": "answer_execution",
  "payload_variant": "compact",
  "query": "{user query}",
  "graph": "{revised graph}",
  "execution_wave": {
    "index": 1,
    "target_node_ids": ["N7", "N8"]
  },
  "shared_dependencies": [
    {
      "parent_id": "N1",
      "parent_type": "search",
      "provision": "inline|history_reference",
      "available_citation_ids": ["N1-R1", "N1-F1"],
      "evidence_items": "{included only when provision=inline}"
    }
  ],
  "target_nodes": [
    {
      "id": "N7",
      "type": "aggregate",
      "need": "{node need}",
      "key_points": ["{point one}", "{point two}"],
      "dependency_refs": [
        {
          "parent_id": "N1",
          "parent_type": "search",
          "edge_type": "enables",
          "instruction": "Use the following as direct supporting evidence:"
        }
      ]
    }
  ],
  "instruction": "Complete every target aggregate node in this wave..."
}
\end{promptbox}
\promptfigcaption{Answerer aggregate-wave user payload}

\begin{promptbox}
{
  "phase": "answer_final",
  "query": "{user query}",
  "graph": "{revised graph}",
  "target_node": {
    "id": "N11",
    "type": "answer",
    "need": "{final answer need}"
  },
  "branch_reports": [
    {
      "node_id": "N7",
      "need": "{upstream branch need}",
      "report_snippet": "{short branch-report preview}",
      "citation_ids_present": ["N1-R1", "N3-F1"],
      "full_report_available_in_conversation": true
    }
  ],
  "instruction": "Write the terminal answer node using only the provided branch reports. Start with a very brief `<think>` block, then emit exactly one `<node id=\"...\">...</node>` block for the final answer node and nothing else. Integrate the branches into a coherent, polished answer without re-opening retrieval or adding unsupported facts. For deep-research queries, default to a long, detailed answer with clear sections, explicit synthesis, and sustained reasoning rather than a short summary. Preserve and reuse the inline evidence citation ids already established in the upstream branch reports."
}
\end{promptbox}
\promptfigcaption{Answerer final-answer user payload}

\paragraph{SFT trajectory generation.}

The same planner and answerer prompts above are used for cold-start SFT
trajectory generation. Teacher rollouts follow the production sequence:
the planner first receives the raw query and emits an initial graph;
the environment executes the initial search nodes; the planner then
receives the after-search payload, revises the graph, and selects URLs
to fetch; the environment executes any newly added searches and fetches
the selected pages; and the answerer executes aggregate waves and the
terminal answer node using the answerer prompts above. The turn-level
SFT corpus stores the resulting planner and answerer chat turns as
prompt/completion examples.

\subsection{Judge and Reward Prompts}

The answerer synthesis reward uses two terms: the LLM final-answer rubric judge below and deterministic branch utilization. Branch utilization is computed as the mean fraction of each aggregate branch's content tokens that appear in the final answer, after stripping inline citations; it does not call an LLM judge and has no prompt.

\begin{promptbox}
You create evaluation rubrics for deep research questions.

Your job is to write 8-10 distinct rubric items that together capture what a strong deep-research answer must do.

Requirements:
- Rubrics must be specific to the query, not generic boilerplate.
- Rubrics should be non-overlapping as far as possible.
- Rubrics should cover substantive facets, evidence quality, synthesis quality, caveats, and practical usefulness when relevant.
- Each rubric needs:
  - `title`: short label
  - `description`: one precise sentence explaining what must be covered
- Avoid style-only rubrics unless structure is essential for the query.

Return JSON only:
{
  "rubrics": [
    {
      "title": "Facet name",
      "description": "What a strong answer must cover."
    }
  ]
}
\end{promptbox}
\promptfigcaption{Rubric generation system prompt}

\begin{promptbox}
You are grading a deep research answer against query-specific rubrics.

You will receive:
- the user query
- the final answer
- a list of rubrics

Score each rubric independently using exactly one of these five scores:
- `0`: not covered -- the rubric is absent, wrong, or unsupported
- `0.25`: minimally covered -- the rubric is touched on but mostly inadequate
- `0.5`: partially covered -- the rubric is addressed but with notable gaps or weak support
- `0.75`: mostly covered -- the rubric is substantively addressed with minor gaps
- `1`: fully covered -- the rubric is clearly and completely satisfied

Rules:
- judge each rubric independently
- score substance, specificity, and support, not writing style
- do not reward confident language if the answer is vague or unsupported
- use the full five-level range when distinctions are meaningful
- use `0.25` when the answer only gestures at a rubric without useful substance
- use `0.5` when the answer addresses the rubric but leaves notable gaps
- use `0.75` when the answer covers the rubric well but has minor gaps or weak support
- use `1` only when the rubric is clearly satisfied with specific, well-supported content

Return JSON only:
{
  "judgements": [
    {
      "rubric_id": "R1",
      "score": 0
    }
  ]
}
\end{promptbox}
\promptfigcaption{Answer judge system prompt}

\clearpage

\section{Example Planner--Answerer Rollout}
\label{app:example_rollout}

\begin{promptbox}
How does nitrogen uptake by tree roots in agroforestry intercropping systems help reduce subsoil nitrogen leaching?
\end{promptbox}
\promptfigcaption{User query}

\paragraph{Abbreviation policy.}

This example keeps the control flow and graph structure intact while
excerpting long evidence payloads and repetitive answer prose. The
important structural property is that the revised graph has two
aggregate execution waves before the final answer: N7--N9 are executed
first, N10 is executed second, and the final answer consumes N7, N8,
and N10. N9 therefore shapes the final report indirectly through N10
rather than as a direct parent of the answer node.

\paragraph{Turn Flow.}

\begin{longtable}{p{0.16\linewidth}p{0.25\linewidth}p{0.51\linewidth}}
\toprule
Turn & Model / environment action & Nodes or artifacts \\
\midrule
\endhead
Planner turn 1 & Emit initial typed DAG & Search nodes N1--N5, aggregate nodes N6--N8, answer node N9. \\
Environment & Execute initial searches with Jina Search & Search results attached to N1--N5. \\
Planner turn 2 & Revise graph and request fetches & Adds search node N6, revises aggregate nodes N7--N10, answer node N11, and selects four fetches. \\
Environment & Execute new search and fetch selected pages with Serper & New search: N6. Fetched evidence: N3-F1, N3-F2, N3-F3, N5-F4. \\
Answerer wave 1 & Execute ready aggregate nodes & N7: tree-root uptake and retention; N8: agroforestry nitrogen cycling; N9: preferential flow and soil transport. \\
Answerer wave 2 & Execute dependent aggregate node & N10: evidence that subsoil nitrogen leaching is reduced by tree-root systems. \\
Answerer final & Write terminal answer node & N11 integrated answer using N7, N8, and N10. \\
\bottomrule
\caption{Turn flow for the agroforestry nitrogen case-study rollout.}
\label{tab:appendix_rollout_turn_flow}\\
\end{longtable}

\paragraph{Revised Graph Summary.}

\begin{longtable}{p{0.08\linewidth}p{0.12\linewidth}p{0.51\linewidth}p{0.20\linewidth}}
\toprule
Node & Type & Need & Inputs \\
\midrule
\endhead
N1 & search & Mechanisms by which tree roots take up nitrogen and reduce nitrogen loss & -- \\
N2 & search & Agroforestry intercropping systems and how they influence nitrogen cycling & -- \\
N3 & search & Evidence on subsoil nitrogen leaching reduction in agroforestry or tree-root systems & -- \\
N4 & search & Role of root architecture, depth, and temporal dynamics in nitrogen capture & -- \\
N5 & search & Soil water and nutrient transport processes affecting leaching & -- \\
N6 & search & Direct evidence on preferential flow and how tree roots alter leaching pathways & -- \\
N7 & aggregate & How tree roots reduce nitrogen loss through uptake and retention & N1, N4 \\
N8 & aggregate & How agroforestry intercropping changes nitrogen cycling and belowground environment & N2, N5 \\
N9 & aggregate & How preferential flow and soil transport processes influence leaching & N5, N6 \\
N10 & aggregate & Evidence that subsoil nitrogen leaching is reduced by tree-root systems & N3, N5, N9 \\
N11 & answer & Integrated explanation of how nitrogen uptake by tree roots in agroforestry intercropping reduces subsoil nitrogen leaching & N7, N8, N10 \\
\bottomrule
\caption{Revised graph summary for the agroforestry nitrogen rollout.}
\label{tab:appendix_rollout_graph_summary}\\
\end{longtable}

\begin{promptbox}
<think>
I'll break this into mechanism, system-specific factors, and evidence branches, then synthesize them into one final answer.
</think>
{
  "graph": {
    "nodes": [
      {"id": "N1", "type": "search", "need": "Mechanisms by which tree roots take up nitrogen and reduce nitrogen loss", "query": "tree root nitrogen uptake mechanisms nitrate ammonium root absorption mycorrhizae subsoil nitrogen retention"},
      {"id": "N2", "type": "search", "need": "Agroforestry intercropping systems and how they influence nitrogen cycling", "query": "agroforestry intercropping nitrogen cycling legumes trees root biomass belowground dynamics"},
      {"id": "N3", "type": "search", "need": "Evidence on subsoil nitrogen leaching reduction in agroforestry or tree-root systems", "query": "subsoil nitrogen leaching reduction agroforestry tree roots evidence review"},
      ...
      {"id": "N9", "type": "answer", "need": "Integrated explanation of how nitrogen uptake by tree roots in agroforestry intercropping reduces subsoil nitrogen leaching", "inputs": ["N6", "N7", "N8"]}
    ],
    "edges": [
      {"source": "N1", "target": "N6", "type": "enables"},
      {"source": "N5", "target": "N8", "type": "contextualizes"},
      ...
    ]
  }
}
\end{promptbox}
\promptfigcaption{Planner initial output (abbreviated)}

\begin{figure}[!htbp]
\centering
\includegraphics[width=0.72\linewidth]{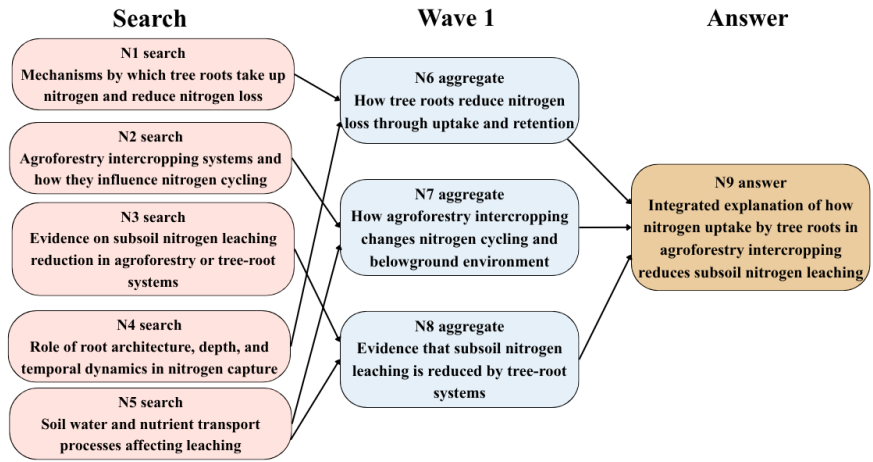}
\caption{Initial agroforestry nitrogen planner graph before search-result revision. Red
nodes are search nodes, blue nodes are aggregate nodes, and the orange node is the
terminal answer node.}
\label{fig:appendix_rollout_graph_pre}
\end{figure}
\FloatBarrier

\begin{promptbox}
<think>
I'll keep the overall structure but sharpen it around the evidence: add a separate branch for preferential flow and leaching pathways, and fetch the most authoritative direct sources.
</think>
{
  "action": "revise",
  "revised_graph": {
    "nodes": [
      {"id": "N1", "type": "search", "need": "Mechanisms by which tree roots take up nitrogen and reduce nitrogen loss", "query": "tree root nitrogen uptake mechanisms nitrate ammonium root absorption mycorrhizae subsoil nitrogen retention"},
      ...
      {"id": "N11", "type": "answer", "need": "Integrated explanation of how nitrogen uptake by tree roots in agroforestry intercropping reduces subsoil nitrogen leaching", "inputs": ["N7", "N8", "N10"]}
    ],
    "edges": [
      {"source": "N5", "target": "N9", "type": "enables"},
      {"source": "N9", "target": "N10", "type": "enables"},
      {"source": "N10", "target": "N11", "type": "enables"},
      ...
    ]
  },
  "fetch_requests": [
    {"source_node_id": "N3", "url": "https://link.springer.com/article/10.1007/s13593-022-00791-7", "reason": "High-level review with direct evidence on reduced N loss in agroforestry."},
    {"source_node_id": "N3", "url": "https://www.nature.com/articles/s43016-022-00657-x", "reason": "Specific evidence that deep tree rooting can catch subsoil inorganic N leached below cropland rooting zones."},
    {"source_node_id": "N3", "url": "https://www.researchgate.net/publication/225134004_Reduced_soil_nutrient_leaching_following_the_establishment_of_tree-based_intercropping_systems_in_eastern_Canada", "reason": "Direct study with quantified subsoil NO3 leaching reduction in tree-based intercropping."},
    {"source_node_id": "N5", "url": "https://www.sciencedirect.com/science/article/abs/pii/S0167198725001278", "reason": "Core source on preferential flow and transport pathways."}
  ]
}
\end{promptbox}
\promptfigcaption{Planner revision output (abbreviated)}

\begin{figure}[!htbp]
\centering
\includegraphics[width=0.82\linewidth]{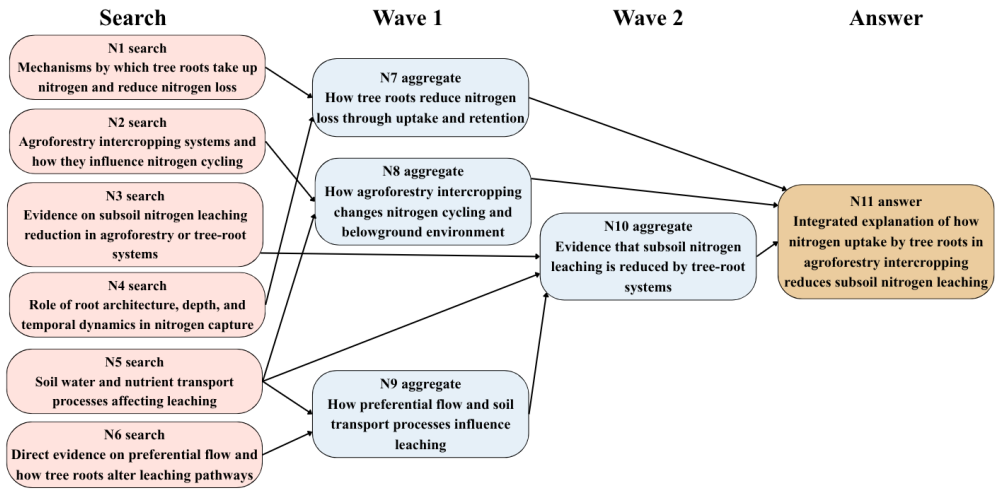}
\caption{Revised graph after search-result revision.}
\label{fig:appendix_rollout_graph_post}
\end{figure}
\FloatBarrier

\begin{promptbox}
{
  "phase": "answer_execution",
  "payload_variant": "compact",
  "execution_wave": {"index": 1, "target_node_ids": ["N7", "N8", "N9"]},
  "shared_dependencies": [
    {
      "parent_id": "N1",
      "parent_type": "search",
      "provision": "inline",
      "available_citation_ids": ["N1-R1", "N1-R3", "N1-R4"],
      "evidence_items": "[N1-R4] Nitrogen Journey in Plants... [search text truncated after excerpt] ..."
    },
    {
      "parent_id": "N5",
      "parent_type": "search",
      "provision": "inline",
      "available_citation_ids": ["N5-R1", "N5-R2", "N5-F4"],
      "evidence_items": "[N5-F4] Characterization of preferential flow and transport pathways... [page text truncated after excerpt] ..."
    },
    ...
  ],
  "target_nodes": [
    {"id": "N7", "dependency_refs": [{"parent_id": "N1"}, {"parent_id": "N4"}]},
    {"id": "N8", "dependency_refs": [{"parent_id": "N2"}, {"parent_id": "N5"}]},
    {"id": "N9", "dependency_refs": [{"parent_id": "N5"}, {"parent_id": "N6"}]}
  ]
}
\end{promptbox}
\promptfigcaption{Answerer execution payload excerpt (abbreviated)}

\begin{promptbox}
<think>
I will synthesize N7, N8, and N9 separately, using only the provided dependencies and keeping each branch focused.
</think>

<node id="N7">
Tree roots reduce nitrogen loss through uptake and retention by capturing different nitrogen forms and by structuring the soil in ways that intercept or store nitrogen before it can leach. The evidence shows that tree roots assimilate nitrate, ammonium, and organic nitrogen sources, including available amino acids [N1-R4]. ...
</node>

<node id="N8">
Agroforestry intercropping changes nitrogen cycling and the belowground environment by altering tree-root biomass, litter inputs, and the balance of competition and facilitation between crops and trees. Leguminous trees transfer organic compounds such as leaf litter, dead roots, and nodules to the soil, enriching the belowground nitrogen pool [N2-R4]. ...
</node>

<node id="N9">
Preferential flow and soil transport processes influence leaching by enabling rapid nutrient transport through macropores and by reducing the time solutes spend in the soil matrix. Preferential flow can bypass the soil matrix and move water and chemical substances over long distances [N5-F4]. ...
</node>
\end{promptbox}
\promptfigcaption{Answerer turn 1 excerpt: answer\_execution (N7, N8, N9)}

\begin{promptbox}
<think>
I'll synthesize the evidence for reduced subsoil nitrogen leaching using the provided direct studies and upstream transport branch.
</think>

<node id="N10">
The evidence base supports the claim that tree-root systems can reduce subsoil nitrogen leaching, but the strength of the evidence varies by study type and context. A review states that agroforestry practices almost consistently reduce nitrogen loss via erosion, runoff, and leaching [N3-F1]. A Nature Food review adds that deep tree rooting can catch and recycle subsoil inorganic nitrogen leached below crop rooting zones [N3-F2]. ...
</node>
\end{promptbox}
\promptfigcaption{Answerer turn 2 excerpt: answer\_execution (N10)}

\begin{promptbox}
<think>
I will integrate the three branch reports into one coherent explanation, preserving their evidence and distinctions.
</think>

<node id="N11">
## Integrated explanation: how nitrogen uptake by tree roots in agroforestry intercropping reduces subsoil nitrogen leaching

Tree roots in agroforestry intercropping systems reduce subsoil nitrogen leaching by capturing nitrogen directly from the soil, structuring the soil to intercept or retain nitrogen before it moves downward, and creating a belowground environment that alters nitrogen cycling and transport pathways. The evidence from the branch reports supports a multi-layered explanation: nitrogen is taken up in different forms, root architecture and depth influence where nitrogen is captured, and belowground interactions in intercropping systems change how nitrogen moves through the soil. At the same time, the evidence base shows that this effect is strongest under certain conditions, while exceptions and limitations exist.

---

## 1) Tree roots capture nitrogen directly and intercept it before it can leach

Tree roots reduce nitrogen loss by absorbing nitrogen from the soil, and the branch reports indicate that they take up several nitrogen forms. The evidence shows that tree roots assimilate nitrogen in the form of nitrate (NO3-), ammonium (NH4+), and organic nitrogen sources, including available amino acids [N1-R4]. A separate source also states that ammonium uptake is mediated by membrane proteins called ammonium transporters (AMTs) [N1-R3]. This means tree roots are not limited to one nitrogen form; they can respond dynamically to the dominant soil pool, but ammonium and nitrate are explicitly highlighted as major assimilation substrates [N1-R3][N1-R4].

Root architecture and depth are central to interception. The branch report says tree roots can structure the soil in ways that intercept or store nitrogen before it can leach, and it cites evidence that root architecture and depth influence nitrogen interception [N4-R1][N4-R4]. Another source in the evidence base notes that root distribution is shaped by nutrient availability and soil conditions, and that root depth is influenced by axial root production [N5-F4]. This implies that deeper root systems are better positioned to capture nitrogen in the subsoil, while shallow roots may intercept nitrogen released by decomposing litter and organic matter [N4-R1][N4-R4].

The branch report also notes that root activity and associated belowground interactions enhance nitrogen retention [N2-R1][N2-R4]. While the evidence does not directly spell out mycorrhizal associations in the excerpted material, the broader context of nitrogen cycling in agroforestry systems implies that tree-root systems can extend nutrient uptake capacity and improve soil nitrogen retention through active root function and associated soil interactions [N1-R1][N1-R3][N2-R1][N2-R4]. In other words, tree roots do not simply absorb nitrogen once; they create conditions in which nitrogen is retained in the soil matrix or taken up before it can leach downward.

---

## 2) Agroforestry intercropping changes nitrogen cycling and the belowground environment

Agroforestry intercropping systems change nitrogen cycling by altering tree-root biomass, litter inputs, and the balance of competition and facilitation between crops and trees. The branch report states that agroforestry systems transfer organic compounds such as leaf litter, dead roots, and nodules to the soil, which enriches the belowground nitrogen pool [N2-R4]. This directly supports the idea that tree-root biomass and litter inputs increase soil nitrogen availability by adding organic nitrogen sources and creating a more nitrogen-rich belowground environment [N2-R4].

The branch report also says that belowground dynamics influence nitrogen cycling and crop productivity, with fine root nitrogen positively correlated to aboveground biomass [N2-R1]. This suggests that tree-root biomass itself contributes to nitrogen cycling by maintaining a belowground nitrogen sink and by influencing the availability of nitrogen for neighboring plants. The intercropping literature further shows that pruning tree canopies changes tree root spatial and temporal distribution, which allows annual crops to develop between the trees [N2-R2]. This pruning effect implies that tree-root distribution is not static; it can shift over time, altering the spatial organization of nitrogen availability and uptake [N2-R2].

Competition and facilitation between crops and trees alter nitrogen dynamics in several ways. The branch report notes that pruning changes root distribution and that root dynamics and belowground interactions are important in agroforestry systems [N2-R2][N2-R3]. This indicates that when trees are present, crops may compete for belowground resources, but pruning can shift the balance toward facilitation by redistributing root space and reducing direct competition [N2-R2][N2-R3]. The evidence also implies that intercropping systems create feedbacks through root biomass, litter inputs, and altered root distribution, which jointly shape nitrogen availability and cycling [N2-R1][N2-R4].

Belowground feedbacks in intercropping systems emerge through root biomass, litter inputs, and altered root distribution. The branch report says these feedbacks are important and that agroforestry systems can change how nitrogen is partitioned among crops and trees [N2-R1][N2-R2][N2-R3][N2-R4]. In other words, the presence of trees does not simply add nitrogen; it reorganizes the soil nitrogen pool and the way crops access nitrogen, often improving retention and reducing losses [N2-R1][N2-R4].

---

## 3) Preferential flow and soil transport processes can either accelerate leaching or be intercepted by tree roots

The branch reports also show that soil water and nutrient transport processes are central to leaching. The evidence indicates that preferential flow is rapid movement bypassing the soil matrix, and that macropores serve as the main or sole channels for long-distance transport of water and chemical substances [N5-F4]. Another source defines preferential flow as non-equilibrium flow under various environmental conditions and emphasizes that it can promote rapid solution leaching below the root zone [N5-R2]. This means nitrogen is not evenly distributed through the soil; instead, it can move quickly through preferential pathways, especially when macropores are present [N5-R1][N5-R2][N5-F4].

Tree roots can intercept or redirect preferential flow by creating root pathways and macropores. The evidence notes that biological activities, including roots and soil fauna, form macropores in farmland [N5-F4]. It also states that trees can enhance preferential flow through macropores formed by their roots and associated soil fauna [N5-F4]. At the same time, the evidence suggests that root distribution and pruning can change root spatial and temporal distribution, which can influence how water and nutrients move through the soil profile [N2-R2][N5-F4]. This implies that tree roots do not passively allow preferential flow; they actively shape it by creating the physical pathways and altering the spatial arrangement of soil structure [N5-F4].

Soil structure and root distribution reduce leaching risk by modifying preferential flow connectivity and tortuosity. The evidence highlights that macropore structure strongly correlates with preferential flow and that preferential flow is affected by the connectivity and tortuosity of macropores [N5-F4]. It also notes that high macroporosity corresponds to a lower stained area ratio in tracer experiments, meaning that preferential flow is more localized and less widespread under certain conditions [N5-F4]. In addition, root architecture and distribution influence how nitrogen is intercepted and how water moves through the soil profile [N4-R1][N4-R4]. Taken together, these findings support the conclusion that soil structure and root distribution reduce leaching risk by limiting the spread of preferential flow and by increasing the time nitrogen spends in the soil matrix before reaching deeper layers [N5-F4][N5-R1][N5-R2].

---

## 4) Evidence that subsoil nitrogen leaching is reduced by tree-root systems

The evidence base supports the claim that tree-root systems can reduce subsoil nitrogen leaching, but the strength of the evidence varies by study type and context. The most direct evidence comes from a meta-analytic review stating that agroforestry practices almost consistently show reductions in nitrogen loss via soil erosion, runoff, and leaching [N3-F1]. A second major review explicitly says that agroforestry systems can increase nitrate retention and mitigate the global impact of a leaky nitrogen cycle in croplands, with deep tree rooting able to catch and recycle subsoil inorganic nitrogen leached below the rooting zone of linked croplands [N3-F2]. This is the clearest evidence that subsoil nitrogen leaching is reduced by tree-root systems.

A third direct study reports that tree roots in a tree-based intercropping system established on clay loam soil decreased subsoil NO3- leaching by 227 kg N ha-1 and 30 kg N ha-1 over two consecutive years [N3-R5]. Taken together, these sources show that reduced subsoil nitrogen leaching is not just a theoretical possibility but a documented pattern under trees and tree-rooted systems [N3-F1][N3-F2][N3-R5].

The evidence also suggests that the effect is amplified under certain environmental and management conditions. The review on agroforestry notes that N loss reductions occur through soil erosion, runoff, and leaching, implying that the reduction is most visible when water and nutrient transport pathways are altered [N3-F1]. The subsoil nitrogen review emphasizes that deep tree rooting can catch nitrogen leached below the rooting zone of linked croplands, which suggests that the benefit is especially strong where there is a nearby cropland source of leached nitrogen and where trees extend their roots below that zone [N3-F2]. The direct intercropping study is set in clay loam soil, indicating that soil texture can matter for root development and leaching control [N3-R5]. The preferential flow background also shows that macropores and soil structure strongly influence preferential flow, and that tree roots can create or alter those structures [N5-F4]. This implies that root distribution, soil texture, and the presence of macropores can amplify the leaching-reduction effect [N5-F4].

However, the evidence base has important limitations and exceptions. First, the meta-analytic review notes that N2O and NO emissions vary across agroforestry practices, and that NO emissions are similar to monocropping in some contexts, so nitrogen management is not universally straightforward [N3-F1]. Second, the global synthesis on soil gross N cycling notes that ecosystem N cycling patterns are highly variable and that no single general pattern can be applied broadly across terrestrial ecosystems [N3-F2]. This means that the reduction in leaching is not guaranteed everywhere; it depends on local soil type, climate, and land use. Third, the direct intercropping study is specific to eastern Canada and clay loam soil, so its results may not generalize directly to sandy, dry, or highly disturbed soils [N3-R5]. Finally, the broader evidence on preferential flow shows that high macroporosity can reduce the stained area ratio in tracer experiments, but it can also create faster, more localized preferential flow [N5-F4]. This means that tree roots can reduce leaching in some cases, but they may also create faster transport pathways in others, depending on soil structure and root architecture [N5-F4].

---

## 5) Integrated conclusion

Nitrogen uptake by tree roots in agroforestry intercropping systems helps reduce subsoil nitrogen leaching through several mechanisms:

- **Direct uptake**: Tree roots absorb nitrate, ammonium, and organic nitrogen sources, reducing the amount available to leach [N1-R3][N1-R4].
- **Interception and retention**: Deeper root systems and root architecture allow tree roots to intercept nitrogen in the subsoil, while active root activity and belowground interactions enhance nitrogen retention [N4-R1][N4-R4][N2-R1][N2-R4].
- **Belowground environment change**: Tree-root biomass and litter inputs enrich the soil nitrogen pool, while pruning and root distribution alter competition and facilitation between crops and trees, creating feedbacks that improve nitrogen cycling [N2-R4][N2-R1][N2-R2][N2-R3].
- **Interception of preferential flow**: Tree roots can create macropores and alter root distribution, which intercepts or redirects preferential flow and reduces leaching risk [N5-F4][N2-R2].
- **Soil structure and connectivity**: Soil structure and root distribution reduce leaching risk by modifying preferential flow connectivity and tortuosity [N5-F4].

At the same time, the evidence is context-dependent. The most direct studies and reviews show reduced subsoil nitrogen leaching under agroforestry and tree-root systems [N3-F1][N3-F2][N3-R5], but the effect is amplified by deep rooting, nearby cropland sources, and favorable soil conditions, while exceptions exist in sandy, dry, or highly disturbed soils and in systems where preferential flow may create faster transport pathways [N3-F1][N3-F2][N5-F4].

In short, agroforestry intercropping reduces subsoil nitrogen leaching because tree roots do more than just grow downward--they actively capture nitrogen, reshape the soil environment, and intercept preferential flow pathways, all of which make nitrogen less available to leach into deeper layers [N1-R3][N1-R4][N2-R4][N4-R1][N5-F4].
</node>
\end{promptbox}
\promptfigcaption{Answerer final output: answer\_final (N11)}

\paragraph{Citation Catalog Used by the Final Answer.}
\footnotesize
\setlength{\tabcolsep}{3pt}
\begin{longtable}{p{0.07\linewidth}p{0.06\linewidth}p{0.04\linewidth}p{0.34\linewidth}p{0.38\linewidth}}
\toprule
ID & Kind & Node & Title & URL \\
\midrule
\endhead
N1-R3 & search & N1 & Molecular fundamentals of nitrogen uptake and transport in trees & \url{https://academic.oup.com/jxb/article/68/10/2489/3091614} \\
N1-R4 & search & N1 & Nitrogen Journey in Plants: From Uptake to Metabolism, Stress ... & \url{https://pmc.ncbi.nlm.nih.gov/articles/PMC10605003/} \\
N2-R1 & search & N2 & Belowground Dynamics Influence Nitrogen Cycling and Crop ... & \url{https://www.frontiersin.org/journals/sustainable-food-systems/articles/10.3389/fsufs.2021.705577/full} \\
N2-R2 & search & N2 & Nitrogen Cycling and Root Dynamics in an Agroforestry System & \url{https://ui.adsabs.harvard.edu/abs/2024EGUGA..26.4555E/abstract} \\
N2-R3 & search & N2 & Below-ground interactions in dryland agroforestry & \url{https://www.sciencedirect.com/science/article/abs/pii/S0378112798003223} \\
N2-R4 & search & N2 & Nitrogen dynamics in agroforestry systems. A review & \url{https://hal.science/hal-04140609/document} \\
N3-F1 & fetch & N3 & Nitrogen dynamics in agroforestry systems. A review & \url{https://link.springer.com/article/10.1007/s13593-022-00791-7} \\
N3-F2 & fetch & N3 & Expanding agroforestry can increase nitrate retention & \url{https://www.nature.com/articles/s43016-022-00657-x} \\
N3-R5 & search & N3 & Reduced soil nutrient leaching following the establishment of tree ... & \url{https://www.researchgate.net/publication/225134004_Reduced_soil_nutrient_leaching_following_the_establishment_of_tree-based_intercropping_systems_in_eastern_Canada} \\
N4-R1 & search & N4 & Root phenotypes for improved nitrogen capture & \url{https://pmc.ncbi.nlm.nih.gov/articles/PMC11420291/} \\
N4-R4 & search & N4 & What Is Root System Architecture and Why Does It Matter? & \url{https://cid-inc.com/blog/what-is-root-system-architecture-and-why-does-it-matter-in-plants/} \\
N5-F4 & fetch & N5 & Characterization of preferential flow and transport pathways under farmland with different land uses & \url{https://www.sciencedirect.com/science/article/abs/pii/S0167198725001278} \\
N5-R1 & search & N5 & Characterization of preferential flow and transport pathways under ... & \url{https://www.sciencedirect.com/science/article/abs/pii/S0167198725001278} \\
N5-R2 & search & N5 & Characteristics of Preferential Flow Paths and Their Impact on ... & \url{https://www.pjoes.com/pdf-89388-23248?filename=Characteristics-of-Prefer.pdf} \\
\bottomrule
\caption{Citation catalog used by the final answer in the agroforestry nitrogen rollout.}
\label{tab:appendix_rollout_citation_catalog}\\
\end{longtable}
\normalsize

\begin{promptbox}
[Citations]
- [N1-R4] Nitrogen Journey in Plants: From Uptake to Metabolism, Stress ... | https://pmc.ncbi.nlm.nih.gov/articles/PMC10605003/
- [N1-R3] Molecular fundamentals of nitrogen uptake and transport in trees | https://academic.oup.com/jxb/article/68/10/2489/3091614
- [N4-R1] Root phenotypes for improved nitrogen capture - PMC - NIH | https://pmc.ncbi.nlm.nih.gov/articles/PMC11420291/
- [N4-R4] What Is Root System Architecture and Why Does It Matter? | https://cid-inc.com/blog/what-is-root-system-architecture-and-why-does-it-matter-in-plants/
- [N5-F4] Characterization of preferential flow and transport pathways under farmland with different land uses - ScienceDirect | https://www.sciencedirect.com/science/article/abs/pii/S0167198725001278
- [N2-R1] Belowground Dynamics Influence Nitrogen Cycling and Crop ... | https://www.frontiersin.org/journals/sustainable-food-systems/articles/10.3389/fsufs.2021.705577/full
- [N2-R4] Nitrogen dynamics in agroforestry systems. A review - HAL | https://hal.science/hal-04140609/document
- [N2-R2] Nitrogen Cycling and Root Dynamics in an Agroforestry System - ADS | https://ui.adsabs.harvard.edu/abs/2024EGUGA..26.4555E/abstract
- [N2-R3] Below-ground interactions in dryland agroforestry - ScienceDirect | https://www.sciencedirect.com/science/article/abs/pii/S0378112798003223
- [N5-R2] Characteristics of Preferential Flow Paths and Their Impact on ... | https://www.pjoes.com/pdf-89388-23248?filename=Characteristics-of-Prefer.pdf
- [N5-R1] Characterization of preferential flow and transport pathways under ... | https://www.sciencedirect.com/science/article/abs/pii/S0167198725001278
- [N3-F1] Nitrogen dynamics in agroforestry systems. A review | https://link.springer.com/article/10.1007/s13593-022-00791-7
- [N3-F2] Expanding agroforestry can increase nitrate retention and mitigate the global impact of a leaky nitrogen cycle in croplands | https://www.nature.com/articles/s43016-022-00657-x
- [N3-R5] Reduced soil nutrient leaching following the establishment of tree ... | https://www.researchgate.net/publication/225134004_Reduced_soil_nutrient_leaching_following_the_establishment_of_tree-based_intercropping_systems_in_eastern_Canada
\end{promptbox}
\promptfigcaption{Citation block appended to final answer}